\documentclass[10pt,twocolumn,letterpaper]{article}

\usepackage{cvpr}              

\usepackage{graphicx}
\graphicspath{ {images/} }
\usepackage{amsmath}
\usepackage{amssymb}
\usepackage{mathtools}
\usepackage{bm}
\usepackage{booktabs}
\usepackage{lipsum}
\usepackage{multirow}
\usepackage{tabularx}
\usepackage{makecell}
\usepackage{amssymb}
\usepackage{pifont}
\usepackage{adjustbox}
\usepackage[percent]{overpic}

\usepackage{enumitem}
\setlist[itemize]{topsep={0pt},partopsep={0pt}}

\usepackage{caption}
\renewcommand{\arraystretch}{1.1}

\usepackage[dvipsnames]{xcolor}
\usepackage{colortbl}
\definecolor{citecolor}{HTML}{1976D2}
\definecolor{refcolor}{HTML}{8E24AA}
\definecolor{linkcolor}{rgb}{0.956,0.298,0.235} 
\definecolor{fontcolor}{rgb}{0.267,0.420,0.809} 

\definecolor{gray}{gray}{0.95}
\definecolor{cyan}{rgb}{0.831,0.901,0.945}

\usepackage{weva}
\usepackage[pagebackref,breaklinks,colorlinks,linkcolor=linkcolor,citecolor=citecolor]{hyperref}

\usepackage[capitalize]{cleveref}
\crefname{section}{Sec.}{Secs.}
\Crefname{section}{Section}{Sections}
\Crefname{table}{Table}{Tables}
\crefname{table}{Tab.}{Tabs.}

\newcommand{\zb}{{\boldsymbol z}}
\newcommand{\xb}{{\boldsymbol x}}
\newcommand{\epsilonb}{\boldsymbol{\epsilon}}

\def\cvprPaperID{****} 
\def\confName{CVPR}
\def\confYear{2023}

\begin{document}

\title{Rodin: A Generative Model for Sculpting 3D Digital Avatars Using Diffusion}

\author{
Tengfei Wang$^{1\dagger}$\footnotemark[1] 
\quad Bo Zhang$^2$\footnotemark[1]
\quad Ting Zhang$^2$
\quad Shuyang Gu$^2$
\quad Jianmin Bao$^2$\\
\quad Tadas Baltrusaitis$^2$ 
\quad Jingjing Shen$^2$
\quad  Dong Chen$^2$
\quad Fang Wen$^2$
\quad Qifeng Chen$^1$
\quad Baining~Guo$^2$\\
{$^1$HKUST \quad $^2$Microsoft Research} \\
{}
}


\twocolumn[{
\renewcommand\twocolumn[1][]{#1}
\maketitle
\vspace{-0.45cm}
\centering

\vspace{0.02em}
\begin{overpic}
    [width=0.99\textwidth]{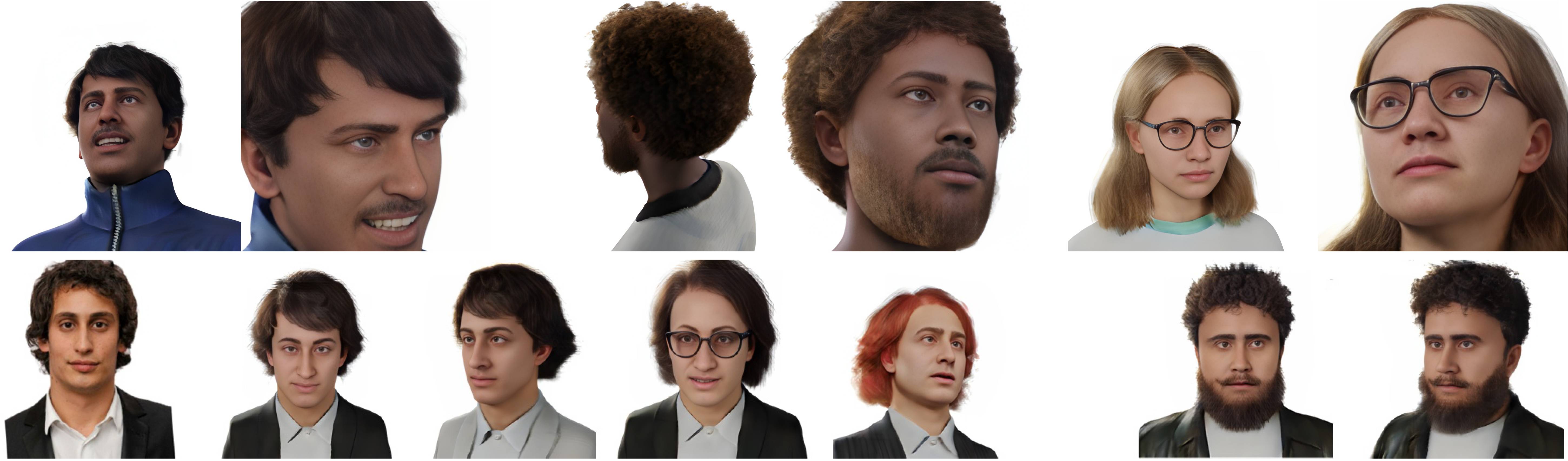}
    \put(2,-1.8){\footnotesize Input portrait}
    \put(13,-1.8){\footnotesize Generated 3D avatar}
    \put(28.2,-1.8){\footnotesize ``\textit{In white suit}''}
    \put(38.8,-1.8){\footnotesize ``\textit{Smile with glasses}''}
    \put(53.5,-1.8){\footnotesize ``\textit{With red hair}''}
    \put(66.8,-1.8){\footnotesize ``\textit{bearded man with curly hair in black leather jacket}''}
  \end{overpic}  
\captionsetup{type=figure}
\vspace{0.9em}
\caption{Our diffusion model, Rodin, can produce high-fidelity 3D avatars,
as shown in the first row. Our model also supports 3D avatar generation from a single portrait or text prompt, while permitting text-based semantic manipulation (second row). See the \href{https://3d-avatar-diffusion.microsoft.com/}{\color{refcolor}{{webpage}}} for video demos.}
\vspace{-0.4cm}
\label{fig:teaser}
\vspace{1cm}
}]

\footnotetext[1]{Equal contribution. $^\dagger$Intern at Microsoft Research.}
\begin{abstract}
\vspace{-0.3cm}
This paper presents a 3D generative model that uses diffusion models to automatically generate 3D digital avatars represented as neural radiance fields. A significant challenge in generating such avatars is that the memory and processing costs in 3D are prohibitive for producing the rich details required for high-quality avatars. To tackle this problem we propose the roll-out diffusion network (Rodin), which represents a neural radiance field as multiple 2D feature maps and rolls out these maps into a single 2D feature plane within which we perform 3D-aware diffusion. The Rodin model brings the much-needed computational efficiency while preserving the integrity of diffusion in 3D by using 3D-aware convolution that attends to projected features in the 2D feature plane according to their original relationship in 3D. We also use latent conditioning to orchestrate the feature generation for global coherence, leading to high-fidelity avatars and enabling their semantic editing based on text prompts. Finally, we use hierarchical synthesis to further enhance details. The 3D avatars generated by our model compare favorably with those produced by existing generative techniques. We can generate highly detailed avatars with realistic hairstyles and facial hair like beards. We also demonstrate 3D avatar generation from image or text as well as text-guided editability. 
\end{abstract}
\vspace{-0.5cm}

\section{Introduction}
\label{sec:intro}

Generative models~\cite{liu2021generative,aggarwal2021generative} are one of the most promising ways to analyze and synthesize visual data including 2D images and 3D models. At the forefront of generative modeling is the diffusion model~\cite{ho2020denoising,song2020score,dhariwal2021diffusion}, which has shown phenomenal generative power for images~\cite{ramesh2022hierarchical,saharia2022photorealistic,rombach2022high,gu2022vector} and videos~\cite{ho2022imagen,singer2022make}. Indeed, we are witnessing a 2D content-creation revolution driven by the rapid advances of diffusion and generative modeling. 

In this paper, we aim to expand the applicability of diffusion such that it can serve as a generative model for 3D digital avatars. We use ``digital avatars'' to refer to the traditional avatars manually created by 3D artists, as opposed to the recently emerging photorealistic avatars~\cite{cao2022authentic,park2021nerfies}. The reason for focusing on digital avatars is twofold. On the one hand, digital avatars are widely used in movies, games, the metaverse, and the 3D industry in general. On the other hand, the available digital avatar data is very scarce as each avatar has to be painstakingly created by a specialized 3D artist using a sophisticated creation pipeline~\cite{guo2019relightables,lombardi2018deep}, especially for modeling hair and facial hair. All this leads to a compelling scenario for generative modeling. 

We present a diffusion model for automatically producing digital avatars represented as neural radiance fields~\cite{mildenhall2021nerf}, with each point describing the color radiance and density of the 3D volume. The core challenge in generating neural volume avatars is the prohibitive memory and computational cost for the rich details required by high-quality avatars. Without rich details, our results will always be somewhat ``toy-like". To tackle this challenge, we develop Rodin, the roll-out diffusion network. We take a neural volume represented as multiple 2D feature maps and roll out these maps into a single 2D feature plane and perform 3D-aware diffusion within this plane. Specifically, we use the tri-plane representation~\cite{chan2022efficient}, which represents a volume by three axis-aligned orthogonal feature planes. By simply rolling out feature maps, the Rodin model can perform 3D-aware diffusion using an efficient 2D architecture and drawing power from the model’s three key ingredients below. 

The first is the 3D-aware convolution. The 2D CNN processing used in conventional 2D diffusion cannot well handle the feature maps originating from orthogonal planes. Rather than treating the features as plain 2D input, the 3D-aware convolution explicitly accounts for the fact that a 2D feature in one plane (of the tri-plane) is a projection from a piece of 3D data and is hence intrinsically associated with the same data’s projected features in the other two planes. To encourage cross-plane communication we involve all these associated features in the convolution and thus bridge the associated features together and synchronize their detail synthesis according to their 3D relationship. 

The second key ingredient is latent conditioning. We use a latent vector to orchestrate the feature generation so that it is globally coherent across the 3D volume, leading to better-quality avatars and enabling their semantic editing. We do this by using the avatars in the training dataset to train an additional image encoder which extracts a semantic latent vector serving as the conditional input to the diffusion model. This latent conditioning essentially acts as an autoencoder in orchestrating the feature generation. For semantic editability, we adopt a frozen CLIP image encoder~\cite{chan2022efficient} that shares the latent space with text prompts. 

The final key ingredient is hierarchical synthesis. We start by generating a low-resolution tri-plane ($64 \times 64$), followed by a diffusion-based upsampling that yields a higher resolution ($256 \times 256$). When training the diffusion upsampler, it is instrumental in penalizing the image-level loss that we compute in a patch-wise manner.

Taken together, the above ingredients work in concert to enable the Rodin model to coherently perform diffusion in 3D with an efficient 2D architecture. The Rodin   is trained with a multi-view image dataset of 100K avatars of diverse identities, hairstyles, and clothing created by 3D artists~\cite{wood2021fake}. 

Several application scenarios are thus supported. We can use the model to generate an unlimited number of avatars from scratch, each avatar being different from others as well as the ones in the training data. As shown in Figure~\ref{fig:teaser}, we can generate highly-detailed avatars with realistic hairstyles and facial hairs styled as beards, mustaches, goatees, and sideburns. Hairstyle and facial hair are essential for representing people’s unique personal identities. Yet, these styles have been notoriously difficult to generate well with existing approaches. The Rodin model also allows avatar customization with the resulting avatar capturing the visual characteristics of the person portrayed in the image or the textual description. Finally, our framework supports text-guided semantic editing. The strong generative power of diffusion shows great promise in 3D modeling.

\begin{figure*}[t]
  \centering
  \small
  \begin{overpic}
    [width=0.98\textwidth]{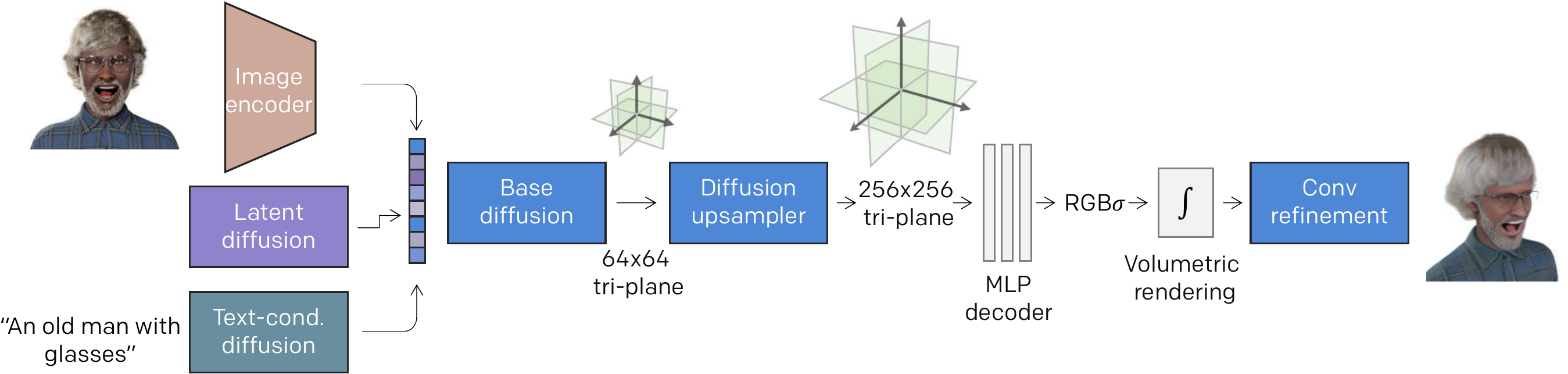}
    \put(24.4,11){\scriptsize $\bm{z}$}
    \put(1.9,8.5){\scriptsize $\bm{z}_0\sim \mathcal{N}(\bm{0},\bm{I})$}
  \end{overpic} 
  \caption{An overview of our Rodin model. We derive the latent $\bm{z}$ via the mapping from image, text, or random noise, which is used to control the base diffusion model to generate $64\times 64$ tri-planes. We train another diffusion model to upsample this coarse result to $256\times 256$ tri-planes that are used to render final multi-view images with volumetric rendering and convolutional refinement. The operators used in diffusion models are designed to be 3D-aware. }
  \label{framework}
  \label{fig:framework}
\end{figure*}

\section{Related Work}
\label{sec:related}

The state of generative modeling~\cite{dhariwal2021diffusion,karras2019style,zhang2022styleswin,ramesh2021zero,vahdat2020nvae,esser2021taming,bond2021deep} has seen rapid progress in past years. Diffusion models~\cite{ho2020denoising,song2020score,dhariwal2021diffusion,yang2022diffusion} have recently shown unprecedented
generative ability and compositional power. The most remarkable success happens in text-to-image
synthesis~\cite{nichol2021glide,ramesh2022hierarchical,gu2022vector,rombach2022high,saharia2022photorealistic}, which serves as a foundation model and enables various appealing applications~\cite{ruiz2022dreambooth,hertz2022prompt,wang2022pretraining} previously unattainable.
While diffusion models have been successfully applied to
different modalities~\cite{ho2022imagen,chen2022analog,li2022diffusion,jeong2021diff}, its generative capability is much less explored in 3D generation, with only a few attempts on modeling 3D primitives~\cite{luo2021diffusion,zhou20213d,zeng2022lion}. 

Early 3D generation works~\cite{shi2022deep} rely on either GAN~\cite{goodfellow2020generative} or VAE~\cite{kingma2019introduction} to model the distribution of 3D shape representation like voxel grids~\cite{wu2016learning,brock2016generative}, point
clouds~\cite{achlioptas2018learning,cai2020learning,xie2021style,li2021sp}, mesh~\cite{szabo2019unsupervised,liao2020towards} and implicit neural representation~\cite{park2019deepsdf,sitzmann2020implicit}. However,
existing works have not demonstrated the ability to produce complex 3D assets yet. Concurrent to this work, Bautista~\etal~\cite{bautista2022gaudi} train a diffusion model to generate the latent vector that encodes the radiance field~\cite{mildenhall2021nerf} of synthetic scenes,
yet this work only produces coarse 3D geometry.  In comparison, we propose a hierarchical 3D generation framework with effective 3D-aware operators, offering unprecedented 3D detail synthesis. 

Another line of work learns 3D-aware generation by utilizing richly available 2D data. 3D-aware
GANs~\cite{chan2021pi,schwarz2020graf,niemeyer2021giraffe,gu2021stylenerf,zhou2021cips,chan2022efficient,deng2022gram,xiang2022gram,sun2022ide,schwarz2022voxgraf,gao2022get3d} recently attract significant research interest, which are trained to produce radiance fields with  image level distribution matching. 
However, these methods suffer from instabilities and mode collapse of GAN training, and it is still challenging to attain authentic avatars that can be viewed from large angles. Concurrently, there are a few attempts to use diffusion models for
the problem. Daniel~\etal~\cite{watson2022novel} proposes to synthesize novel views with a pose-conditioned 2D diffusion model, yet the results are not intrinsically 3D. Ben~\etal~\cite{poole2022dreamfusion} optimizes a radiance field using the supervision from a pretrained text-to-image diffusion model and produces impressive 3D objects of diverse genres. Nonetheless, pretrained 2D generative networks only offer limited 3D knowledge and inevitably lead to blurry 3D results. A high-quality generation framework in 3D space is still highly desired.

\section{Approach}
\label{sec:method}

Unlike prior methods that learn 3D-aware generation from a 2D image collection, we aim to learn the 3D avatar
generation using the multi-view renderings from the Blender synthetic pipeline~\cite{wood2021fake}. Rather than treating the multi-view images of the same subject as individual
training samples, we fit the volumetric neural representation for each avatar, which is used to explain all the observations from different viewpoints. Thereafter we use 
diffusion models to characterize the distribution of these 3D
instances. Our diffusion-based 3D generation is a hierarchical process --- we first utilize a diffusion model to generate the
coarse geometry, followed by a diffusion upsampler for detail synthesis. As illustrated in Figure~\ref{framework}, the whole 3D portrait generation comprises multiple training stages, which we detail in the following subsections.

\subsection{Robust 3D Representation Fitting}
To train a generative network with explicit 3D supervision, we need an expressive 3D representation that accounts for multi-view images, which should meet the following requirements. First, we need an explicit representation that is amenable to generative network processing.
Second, we require a compact representation that is memory efficient; otherwise, it would be too costly to store a myriad of such 3D instances for training. Furthermore, we expect fast representation fitting since hours of optimization as vanilla NeRF~\cite{mildenhall2021nerf} would make it unaffordable to generate abundant 3D training data as required for generative modeling.

Taking these into consideration, we adopt \emph{tri-plane representation} proposed by~\cite{chan2022efficient} to model the neural radiance field of 3D avatars. Specifically, the 3D volume
is factorized into three axis-aligned orthogonal feature planes, denoted by $\bm{y}_{uv},\bm{y}_{wu},\bm{y}_{vw}\in \mathbb{R}^{H\times W \times C}$, each of
which has
spatial resolution of $H\times W$ and number of channel as~$C$. Compared to voxel grids, the tri-plane representation offers a considerably smaller memory
footprint without sacrificing the
expressivity. Hence, rich 3D information is explicitly memorized in the tri-plane features, and one can query the feature of the 3D point $\bm{p}\in \mathbb{R}^3$ by projecting it
onto each plane and aggregating the retrieved features, \ie, $\bm{y}_{\bm{p}}=\bm{y}_{uv}(\bm{p}_{uv})+\bm{y}_{wu}(\bm{p}_{wu})+\bm{y}_{vw}(\bm{p}_{vw})$. With such positional feature, one can derive the density $\sigma \in \mathbb{R}^+$ and view-dependent
 color $\bm{c}\in \mathbb{R}^3$ of each 3D location given the viewing direction $\bm{d} \in \mathbb{S}^2$ with a lightweight MLP decoder $\mathcal{G}_{\theta}^{\textnormal{MLP}}$, which can be formulated as
\begin{equation}
  \bm{c}(\bm{p},\bm{d}),\sigma(\bm{p}) = \mathcal{G}_{\theta}^{\textnormal{MLP}} \big(\bm{y}_{\bm{p}}, \xi(\bm{y}_{\bm{p}}),\bm{d}\big).
\end{equation}
Here, we apply the Fourier embedding operator $\xi(\cdot)$~\cite{tancik2020fourier} on the queried feature rather than the spatial coordinate. 
The tri-plane features and the MLP decoder are optimized such that the rendering of the neural radiance field matches the multi-view images $\{\bm{x}\}_{N_v}$ for the given subject, where $\bm{x}\in \mathbb{R}^{H_0\times W_0 \times 3}$. We enforce the rendered image given by volumetric rendering~\cite{max1995optical}, \ie, $\hat{\bm{x}}=\mathcal{R}(\bm{c},\sigma)$, to match the corresponding ground truth with mean squared error loss.
Besides, we introduce sparse, smooth, and compact regularizers to reduce the ``floating'' artifacts~\cite{barron2022mip} in free space. For more tri-plane fitting details, please refer to the Appendix.  

\begin{figure}[t]
  \centering
  \scriptsize
  \setlength\tabcolsep{1pt}
   \renewcommand{\arraystretch}{0.95}
  \begin{tabular}{@{}ccc@{}}   
    \includegraphics[width=0.32\linewidth]{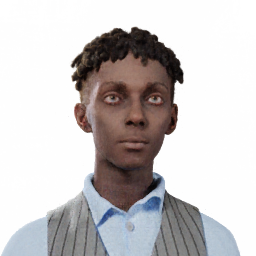}  &
    \includegraphics[width=0.32\linewidth]{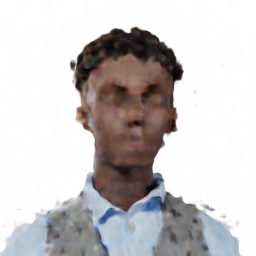}  &    
    \includegraphics[width=0.32\linewidth]{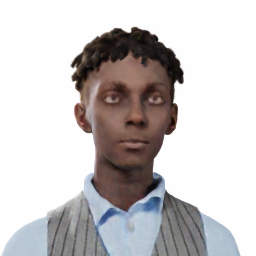} \\        
    [1.4pt]
    (a) $256 \times 256$ tri-plane & (b) $64 \times 64$ tri-plane & (c) $64 \times 64$ tri-plane \\
    [0.5pt]
    &  without random scaling & with random scaling \\
  \end{tabular}
  \caption{While $256\times 256$ tri-planes give good renderings (a), the $64\times 64$ variant gives much worse result (b). Hence, we introduce random scaling during fitting so as to obtain a robust representation that can be effectively rendered in continuous scales (c). }
  \label{fig:triplane-robustness}
\end{figure}

While prior per-scene reconstruction mainly concerns the fitting quality, our 3D fitting procedure should also consider several key aspects
for generation purposes. First, the tri-plane features of different subjects should rigorously reside in the same domain. To achieve this, we adopt a shared MLP decoder when
fitting
distinct portraits, thus implicitly pushing the tri-plane features to the shared latent space recognizable by the decoder. Second, the MLP decoder
has to possess some level of \emph{robustness}. That is, the decoder should be tolerant to slight perturbation of tri-plane features, and thus one can still
obtain plausible results even if the tri-plane features are imperfectly generated. 
More importantly, the decoder should be robust to varied tri-plane sizes because hierarchical 3D generation is trained on multi-resolution tri-plane features. As shown in Figure~\ref{fig:triplane-robustness}, when solely fitting $256\times
256$ tri-planes, its $64\times 64$ resolution variant cannot be effectively rendered. To address this, we randomly scale the tri-plane during fitting,
which is instrumental in deriving multi-resolution tri-plane features simultaneously with a shared decoder.

\subsection{Latent Conditioned 3D Diffusion Model}
\label{sec:latent conditioned}
Now the 3D avatar generation is reduced to learning the distribution of tri-plane features, \ie, $p(\bm{y})$, where $y = (\bm{y}_{uv},\bm{y}_{wu},\bm{y}_{vw})$. Such generative modeling is non-trivial since
$\bm{y}$ is highly dimensional. We leverage diffusion models for the task, which have shown compelling quality in complex image modeling. 

On a high level, the
diffusion model generates $\bm{y}$ by gradually reversing a Markov forward process. Starting from $\bm{y}_0\sim p(\bm{y})$, the forward process $q$ yields a
sequence of increasing noisy latent codes $\{\bm{y}_t \mid t\in [0,T]\}$ according to $\bm{y}_t:=\alpha_t\bm{y}_0 + \sigma_t \bm{\epsilon}$, where $\bm{\epsilon} \in
\mathcal{N}(\bm{0},\bm{I})$ is the added Gaussian noise; 
$\alpha_t$ and $\sigma_t$ define a noise schedule whose log signal-to-noise ratio $\lambda_t = \log[\alpha_t^2/\sigma_t^2] $ linearly decreases with the
timestep $t$.
With sufficient noising steps, we reach a pure Gaussian noise, \ie, $\bm{y}_T\sim \mathcal{N}(\bm{0},\bm{I})$. The generative process corresponds to reversing the
above noising process, where the diffusion model is trained to denoise
$\bm{y}_t$ into $\bm{y}_0$ for all $t$ using a mean squared error loss. Following~\cite{ho2020denoising}, better generation quality can be achieved by parameterizing the diffusion model $\hat{\bm{\epsilon}}_{\theta}$ to
predict the added noise:
\begin{equation}
  \mathcal{L}_{\textnormal {simple }}=\mathbb{E}_{t, \bm{x}_0, \bm{\epsilon}} \Big[\big\|\hat{\bm{\epsilon}}_{\theta}({\alpha_t\bm{y}_0 + \sigma_t \bm{\epsilon}}, t) - \bm{\epsilon}\big\|_2^2 \Big] .
\end{equation}
In practice, our diffusion model training also jointly optimizes the variational lower bound loss $\mathcal{L}_{\textnormal{VLB}}$ as suggested in~\cite{nichol2021improved}, which allows high-quality generation with fewer
timesteps. During inference, stochastic ancestral sampler~\cite{ho2020denoising} is used to generate the final samples, which starts from the Gaussian noise $\bm{y}_T\sim \mathcal{N}(\bm{0},\bm{I})$ and sequentially produces less noisy samples $\{\bm{y}_T,
\bm{y}_{T-1},\ldots\}$ until reaching~$\bm{y}_0$. 
 
\begin{figure}[t]
  \centering
  \small
  \begin{overpic}
    [width=0.96\columnwidth]{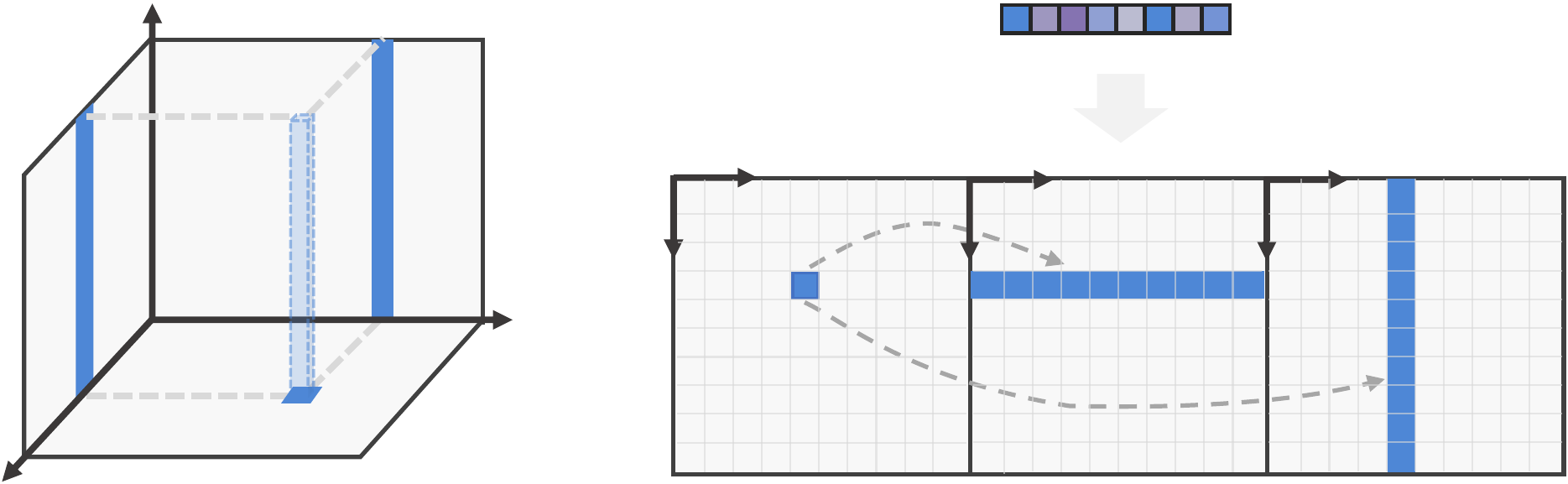}
    \put(-1.95,-2.2){\footnotesize $\bm{u}$}
    \put(33,10){\footnotesize $\bm{v}$}
    \put(8.3,31.2){\footnotesize $\bm{w}$}
    \put(39.5,15.8){\footnotesize $\bm{u}$}
    \put(44.1,21.1){\footnotesize $\bm{v}$}
    \put(58.4,15.8){\footnotesize $\bm{u}$}
    \put(63.1,21.1){\footnotesize $\bm{w}$}
    \put(77.2,15.8){\footnotesize $\bm{w}$}
    \put(82.1,21.1){\footnotesize $\bm{v}$}
    \put(66,32){\footnotesize latent $\bm{z}$}
    \put(15,-5.2){\footnotesize (a)}
    \put(68,-5.2){\footnotesize (b)}
  \end{overpic}
  \vspace{1.4em}
  \caption{We propose two mechanisms to ensure coherent tri-plane generation. Our 3D-aware convolution considers the 3D relationship in (a) and correlates the associated elements from separate feature planes as shown in (b). In (b), we also visualize the usage of a shared latent code to orchestrate the feature generation. }
  \label{fig:3d-aware conv}
\end{figure}

We first train a base diffusion model to generate the coarse-level tri-planes, \eg, at $64\times 64$ resolution. A straightforward approach is to adopt the 2D network structure used in the
state-of-the-art image-based diffusion models for our tri-plane generation. Specifically, we can concatenate the tri-plane features
in the channel dimension as in~\cite{chan2022efficient}, which forms
$\bm{y} = (\bm{y}_{uv}\oplus \bm{y}_{wu}\oplus\bm{y}_{vw})\in \mathbb{R}^{H\times W\times {3C}}$, and employ a well-designed 2D U-Net to model the data distribution through the denoising diffusion process. However, such
a baseline model produces 3D avatars with severe artifacts. 
We conjecture the generation artifact comes from the \emph{incompatibility} between the tri-plane representation and the 2D U-Net. As shown in Figure~\ref{fig:3d-aware conv}(a), intuitively, one can regard the tri-plane features as the projection of neural volume towards the frontal, bottom, and side views, respectively. Hence, the channel-wise concatenation of these orthogonal planes for CNN processing is
problematic because these planes are not spatially aligned. To better handle the tri-plane representation, we make the following efforts.

\noindent\textbf{3D-aware convolution.} Using CNN to process channel-wise concatenated tri-planes will cause the mixing of theoretically uncorrected features in terms of 3D. One simple yet effective way to address this is to spatially \emph{roll out} the tri-plane features. As shown in Figure~\ref{fig:3d-aware conv}(b), we concatenate
the tri-plane features horizontally, yielding
$\tilde{\bm{y}} = \mathtt{hstack}(\bm{y}_{uv},\bm{y}_{wu},\bm{y}_{vw})\in \mathbb{R}^{H\times {3W} \times C}$. Such feature roll-out allows independent processing of feature planes. For simplicity, we subsequently use $\bm{y}$ to denote such input form by default. However, the tri-plane roll-out hampers cross-plane communication, while the 3D generation requires the synergy of the tri-plane generation. 
 
To better process the tri-plane features, we need an efficient 3D operator that performs on the tri-plane rather than treating it as a plain 2D input. To achieve this, we propose \emph{3D-aware convolution} to effectively process the tri-plane features while respecting their 3D relationship. A point on a certain feature plane actually corresponds to an axis-aligned 3D line in the volume, which also has two corresponding line projections in other planes, as shown in Figure~\ref{fig:3d-aware conv}(a). The features of these corresponding locations essentially describe the same 3D primitive and should be learned synchronously. However, such a 3D relationship is neglected when employing plain 2D convolution to tri-plane processing. As such, our 3D-aware convolution explicitly introduces such 3D inductive bias by attending the features of each plane to the corresponding row/column of the rest planes. In this way, we enable 3D processing capability with 2D CNNs. This 3D-aware convolution applied on the tri-plane representation, in fact, is a generic way to simplify 3D convolutions previously too costly to compute when modeling high-resolution 3D volumes.

The 3D-aware convolution is depicted in Figure~\ref{fig:3d-aware conv}(b). Ideally, the compute for $\bm{y}_{uv}$ would attend to full elements from the corresponding row/column, \ie, $\bm{y}_{wu}$ and $\bm{y}_{vw}$, from other planes. For parallel computing, we simplify this and aggregate the row/column elements.   Specifically, we apply the axis-wise pooling for  $\bm{y}_{wu}$ and $\bm{y}_{vw}$, yielding a row vector $\bm{y}_{wu\to u}\in \mathbb{R}^{1\times W \times C}$ and a column vector $\bm{y}_{vw \to v} \in \mathbb{R}^{H\times 1 \times C}$ respectively. For each point of $\bm{y}_{uv}$, we can easily access the corresponding element in the aggregated vectors. We expand the aggregated vectors to the original 2D dimension (\ie, replicating the column vectors along row dimension, and vice versa) and thus derive $\bm{y}_{(\cdot) u}, \bm{y}_{v (\cdot)}\in \mathbb{R}^{H\times W \times C}$. By far, we can perform 2D convolution on the channel-wise
concatenation of the feature maps, \ie, $\mathtt{Conv2D}(\bm{y}_{uv}\oplus \bm{y}_{(\cdot)u}\oplus \bm{y}_{v (\cdot)})$.
because $\bm{y}_{uv}$ is now spatially aligned with the aggregation of the corresponding elements from other planes. 
The compute
for $\bm{y}_{vw}$ and $\bm{y}_{wu}$ is conducted likewise. The 3D-aware convolution greatly enhances the cross-plane
communication, and we empirically observe reduced artifacts and improved generation of thin structures like hair strands.

\noindent\textbf{Latent conditioning.} We further propose to learn a latent vector to \emph{orchestrate} the tri-plane generation. As shown in Figure~\ref{fig:framework}, we additionally train an image
encoder $\mathcal{E}$ to
extract a semantic latent vector serving as the conditional input of the base diffusion model, so essentially the whole framework is an
\emph{autoencoder}. To be specific, we extract the latent vector from the frontal view of each training subject, \ie, $\bm{z}=\mathcal{E}_\theta(\bm{x}_{\textnormal{front}})\in \mathbb{R}^{512}$, and
the diffusion model conditioned on $\bm{z}$ is trained to reconstruct the tri-plane of the same subject. We use adaptive group normalization (AdaGN) to
modulate the activations of the diffusion model, where $\bm{z}$ is injected into
every residual block, and in this way, the features of the orthogonal planes are synchronously generated according to a shared latent. 

The latent conditioning not only leads to higher generation quality but also permits a disentangled latent space, thus allowing semantic
editing of generated results. To achieve better editability, we adopt a frozen CLIP image encoder~\cite{radford2021learning} that has shared latent space with text prompts.
We will show how the learned model produces controllable
text-guided generation results.  

Another notable benefit of latent conditioning is that it allows \emph{classifier-free guidance}~\cite{ho2022classifier}, a technique typically used to boost the sampling quality in the conditional generation.
When training the diffusion model, we randomly zero the latent embedding with $20\%$ probability, thus adapting the
diffusion decoder to unconditional generation. During inference, we
 can steer the model toward better generation sampling according to
\begin{equation}
  \hat{\bm{\epsilon}}_\theta(\bm{y},\bm{z}) = \lambda \bm{\epsilon}_\theta(\bm{y},\bm{z}) + (1-\lambda) \bm{\epsilon}_\theta(\bm{y}),
\end{equation}
where $\bm{\epsilon}_\theta(\bm{y},\bm{z})$ and $\bm{\epsilon}_\theta(\bm{y})$ are the conditional and unconditional $\bm{\epsilon}\text{-}$predictions respectively,
and $\lambda>0$ specifies the guidance strength. 

Our latent conditioned base model thus supports both unconditional generation as well as the conditional generation that is used for portrait inversion. To account for full diversity during unconditional sampling, we additionally train a diffusion model to model the distribution of the latent $\bm{z}$, whereas the latent $\bm{y}_T$ describes the residual variation. We include this latent diffusion model in Figure~\ref{fig:framework}.

\subsection{Diffusion Tri-plane Upsampler}
\label{sec:tri-plane upsampler}
To generate high-fidelity 3D structures, we further train a diffusion super-resolution (SR) model to increase the tri-plane resolution from $64\times 64$ to $256\times 256$. At this stage, the diffusion upsampler 
is conditioned on the low-resolution (LR) tri-plane $\bm{y}^{\textnormal{LR}}$. Different from the base model training we parameterize the diffusion upsampler
${\bm{y}^{\textnormal{HR}}_\theta}(\bm{y}^{\textnormal{HR}}_t, \bm{y}^{\textnormal{LR}}, t)$ to predict
the high-resolution (HR) ground truth $\bm{y}^{\textnormal{HR}}_0$ instead of the added noise $\bm{\epsilon}$. 
The 3D-aware convolution is utilized in each residual block to enhance detail synthesis.

Following prior cascaded image generation works, we apply
\emph{condition augmentation} to reduce the domain gap between the output from the base model and the LR conditional input for SR
training. We conduct careful tuning for the tri-plane augmentation with a combination of random downsampling, Gaussian blurring and Gaussian noises, making the
rendered augmented LR tri-plane resemble the base rendering output  as much as possible.   

\begin{figure*}[t]
  \centering
  \small
  \setlength\tabcolsep{1pt}
  \begin{tabular}{@{}c@{}}
    \includegraphics[width=0.99\linewidth]{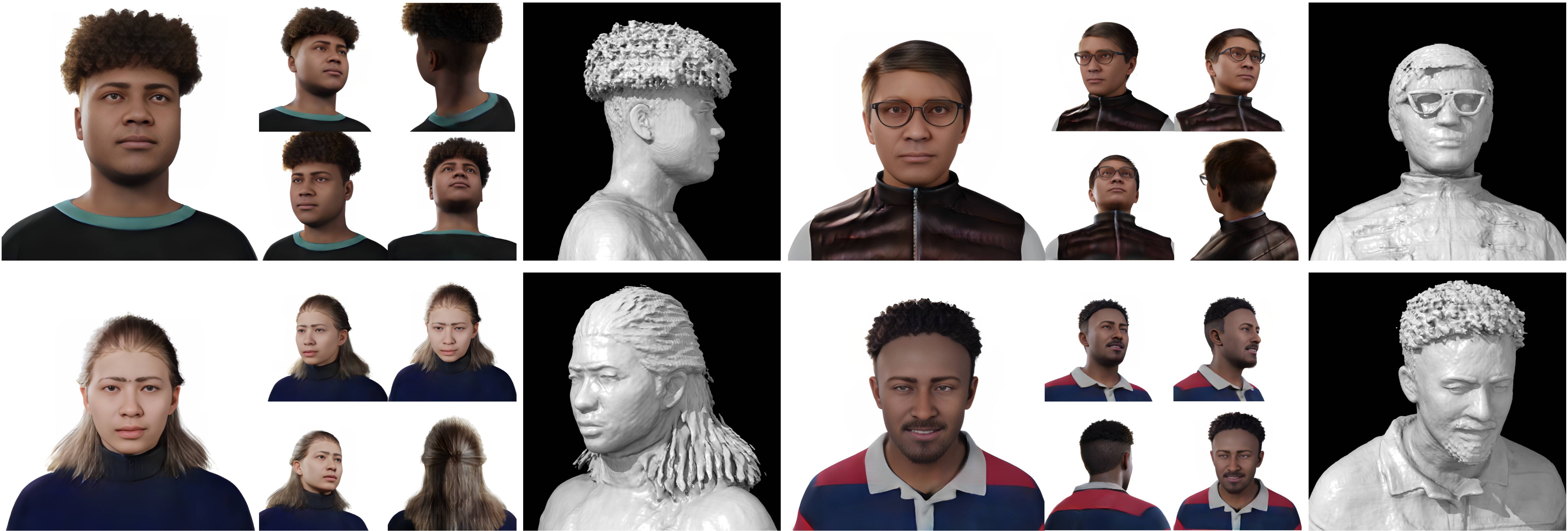}\\
    
  \end{tabular}
  \vspace{-0.5em}
  \caption{Unconditional generation samples by our Rodin model. We visualize the mesh extracted from the generated density field.}
  \label{fig:main-results}
\end{figure*}

\begin{figure}[t]
  \centering
  \small
  \setlength\tabcolsep{1pt}
  \renewcommand{\arraystretch}{0.6}
  \begin{tabular}{@{}c@{}} 
    \includegraphics[width=0.99\linewidth]{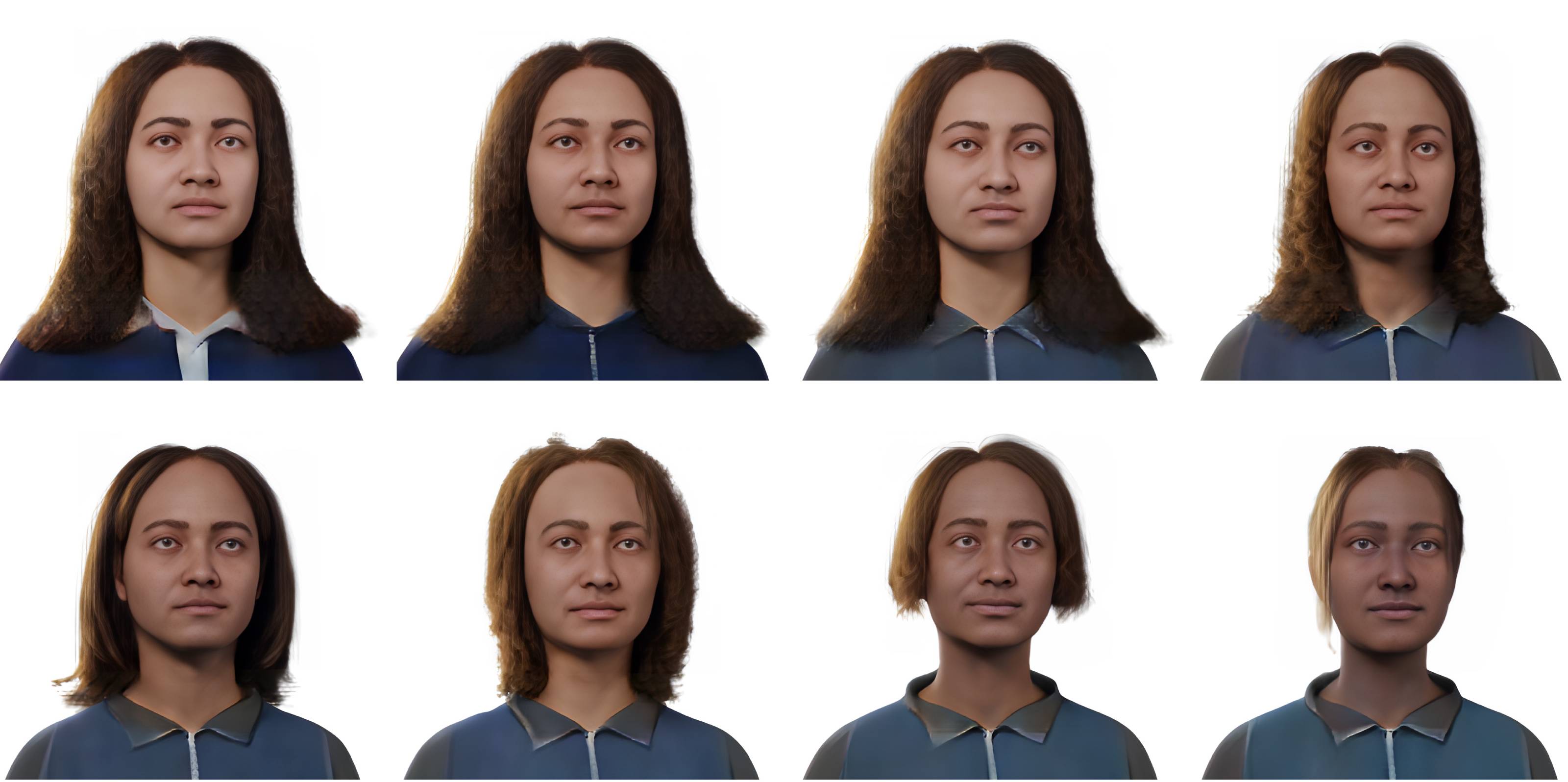} 
     \\       
  \end{tabular}
  \vspace{-0.5em}
  \caption{Latent interpolation results for generated avatars.}
  \label{fig:interpolation}
\end{figure}

Nonetheless, we find that a tri-plane restoration with a lower $\ell_2$ distance to the ground truth may not necessarily correspond to a satisfactory image
rendering. Hence, we need to directly constrain the rendered image. Specifically, we obtain the rendered image
$\hat{\bm{x}^{\textnormal{HR}}}\in \mathbb{R}^{256\times 256\times 3}$ from the
predicted tri-plane $\hat{\bm{y}}^{\textnormal{HR}}_0$ with $ \hat{\bm{x}}=
\mathcal{R}(\mathcal{G}_\theta^{\textnormal{MLP}}(\hat{\bm{y}}^{\textnormal{HR}}_0))$, and we further penalize the perceptual loss~\cite{johnson2016perceptual} 
between this rendered result and the ground truth, according to
\begin{equation}
  \mathcal{L}_{\textnormal {perc}}=\mathbb{E}_{t,\hat{\bm{x}}} \sum_l \lVert \Psi^l(\hat{\bm{x}})-\Psi^l(\bm{x}) \rVert_2^2,
\end{equation}
where $\Psi^l$ denotes the multi-level
feature extraction using a pretrained VGG. Usually, the volume rendering requires stratified sampling along each ray, which is computationally prohibitive
for high-resolution rendering. Hence, we compute $\mathcal{L}_{\textnormal{perc}}$ on random $112\times 112$ image patches with high sampling importance on
face region. Compared with prior 3D-aware GANs that require rendering full images, our 3D-aware SR can be easily scalable to high resolutions due to the permit
of patchwise training with direct supervision.

Modeling high-frequency detail and thin structures are particularly challenging in volumetric rendering. Thus, at this stage, we jointly train a convolution refiner~\cite{wang2021towards} on our data which complements the missing details of the NeRF rendering, ultimately producing compelling $1024\times 1024$ image outputs.

\section{Experiments}
\label{sec:experiment}

\subsection{Implementation Details}
\label{sec:details}
To train our 3D diffusion, we obtain 100K 3D avatars with a random combination of identities, expressions, hairstyles, and accessories using synthetic engine~\cite{wood2021fake}. For each avatar, we render 300 multi-view images with known camera pose, which are sufficient for a high-quality  radiance field reconstruction. 
The tri-planes for our generation have the dimension of $256\times 256\times 32$ in each feature plane.  We optimize a shared MLP decoder  when fitting the first 1,000 subjects. This decoder consists of 4 fully connected layers and is fixed when fitting the following subjects. Thus different subjects are fitted separately in distributed servers. 

Both the base and upsampling diffusion networks
adopt U-Net architecture to process the roll-out tri-plane features. We apply full-attention for $8^2$, $16^2$ and $32^2$ scales within the
network and adopt
3D-aware convolution at higher scales to enhance the details. While we generate $256^2$ tri-planes with the diffusion upsampler, we also render image and compute image loss at $512^2$ resolution, with a convolutional refinement further enhancing the details to $1024^2$. For
more details about the network architecture and training strategies, please refer to our Appendix.

\begin{figure*}[t]
  \centering
  \small
  \setlength\tabcolsep{1pt}
\vspace{0.02em}
\begin{overpic}
    [width=0.99\textwidth]{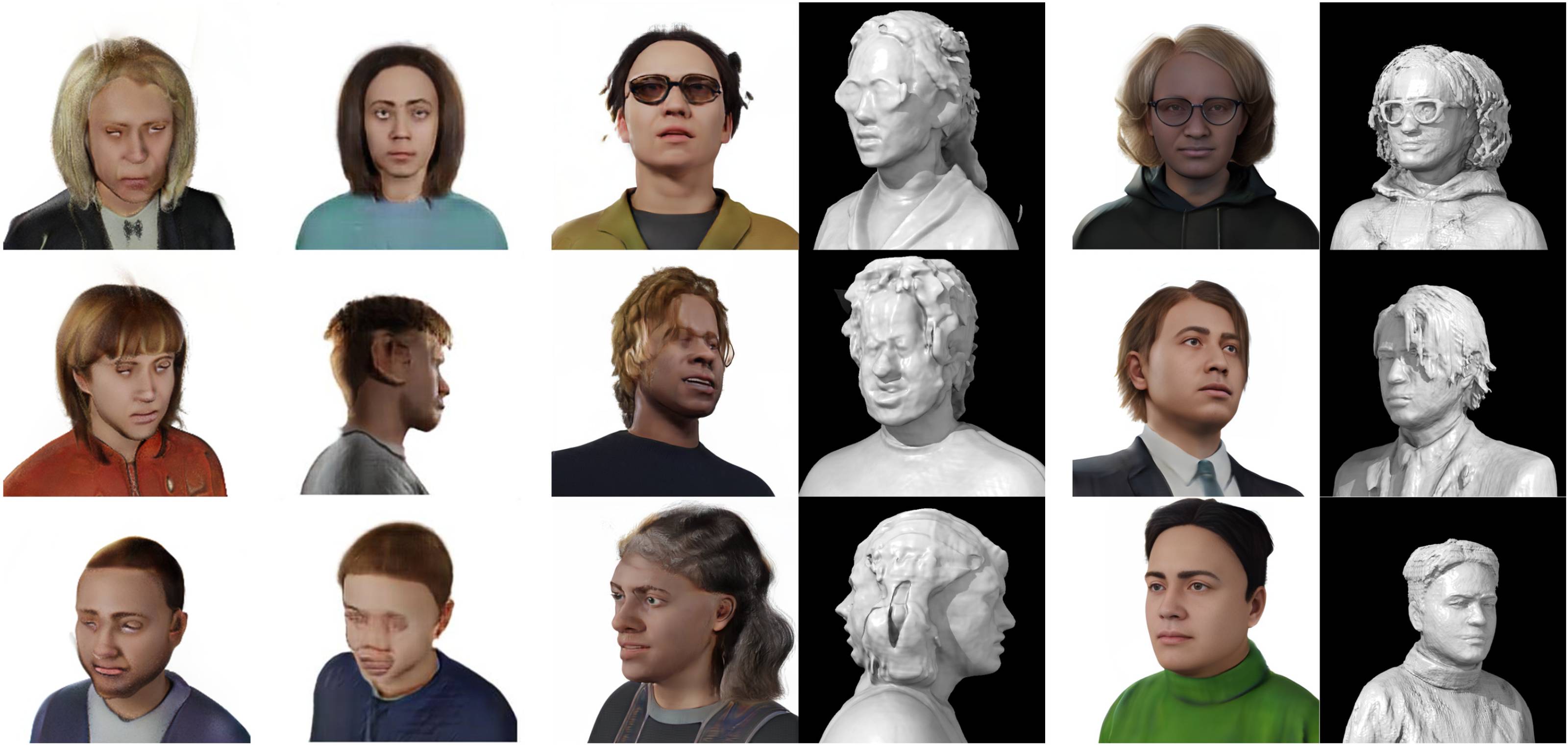}
    \put(5.5,-2){\footnotesize Pi-GAN~\cite{chan2021pi}}
    \put(21.5,-2){\footnotesize GIRAFFE~\cite{niemeyer2021giraffe}}
    \put(48.6,-2){\footnotesize EG3D~\cite{chan2022efficient}}
    \put(79.6,-2){\footnotesize Our Rodin model}
  \end{overpic}  
 
  \vspace{1em}
  \caption{Qualitative comparison with state-of-the-art approaches. }
  \label{fig:comparison}
\end{figure*}

\subsection{Unconditional Generation Results}
\label{sec:results}

\begin{table}[t]
  \centering
  \small
  \begin{tabular}{l@{\hspace{3mm}}c@{\hspace{3mm}}c@{\hspace{3mm}}c@{\hspace{3mm}}c@{\hspace{3mm}}c}
  \toprule
   &  Pi-GAN & GIRAFFE & EG3D & Autoencoder & Ours \\
  \midrule
  FID $\downarrow$ & 78.3& 64.6& 40.5 & 67.4& \textbf{26.1}\\
  \bottomrule
  \end{tabular}
  \caption{\small{Quantitative comparison with baseline methods.}}
  \label{table:quantitative_comparison}
  \end{table} 

\begin{table}[t]
  \centering
  \small
  \begin{tabular}{l@{\hspace{15mm}}c}
  \toprule  
  Model configuration & FID$\downarrow$ \\
  \midrule
  A. Baseline & 39.2\\
  B. + Latent conditioning & 37.4\\
  C. + Tri-plane roll-out & 28.4\\
  D. + 3D-aware conv & \textbf{26.1}\\ 
  \bottomrule
  \end{tabular}
  \caption{Ablation study of the proposed components.}
  \vspace{-5mm}
  \label{table:ablation}
  \end{table}

Figure~\ref{fig:main-results} shows several samples generated by the Rodin model, showing the capability to synthesize high-quality 3D renderings with impressive details, \eg, glasses and hairstyle.  To reflect the geometry, we extract the mesh from the generated density field using marching cubes, which demonstrates high-fidelity geometry. More uncurated samples are shown in the Appendix. We also explore the interpolation of the latent condition $\bm{z}$ between two generated avatars, as shown in Figure~\ref{fig:interpolation}, where we observe consistent high-quality interpolation results with smooth appearance transition.

\subsection{Comparison}
\label{sec:comparison}

We compare our method with state-of-the-art 3D-aware GANs, \eg, Pi-GAN~\cite{chan2021pi} and GIRAFFE~\cite{niemeyer2021giraffe} and EG3D~\cite{chan2022efficient}, which learn to produce neural radiance field from 2D image supervision. Moreover, we implement an auto-encoder baseline, which leverages the multi-view supervision and reconstructs the radiance field from the latent. We differ in this baseline by using the power diffusion-based decoder with 3D-aware designs. We adapt the official implementation of prior works to 360-degree generation and retrain them using the same dataset.

We use FID score~\cite{heusel2017gans} to measure the quality of image renderings. As per~\cite{kynkaanniemi2022role}, we use the features extracted from the CLIP model to compute FID, which we find better correlates the perceptual quality. Specifically, we compute the FID score using 5K generated samples. The quantitative comparison is shown in Table~\ref{table:quantitative_comparison}, where we see that the Rodin model induces significantly lower FID than others.

The visual comparison in Figure~\ref{fig:comparison} shows a clear quality superiority of our Rodin model over prior arts. Our method gives visually pleasing multi-view renderings with high-quality geometry, \eg, for glasses and hair, whereas 3D-aware GANs produce more artifacts due to the geometry ambiguity caused by the simple use of image supervision.

\begin{figure}[t]
  \centering
  \footnotesize
  \setlength\tabcolsep{1pt}
  \renewcommand{\arraystretch}{0.8}
  \vspace{-1em}
  \begin{tabular}{@{}ccc@{}} 
    \includegraphics[width=0.31\linewidth]{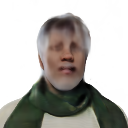}   &
    \includegraphics[width=0.31\linewidth]{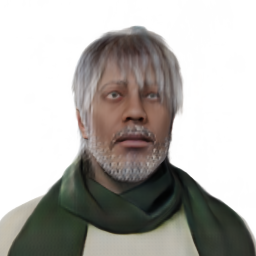}  &
    \includegraphics[width=0.31\linewidth]{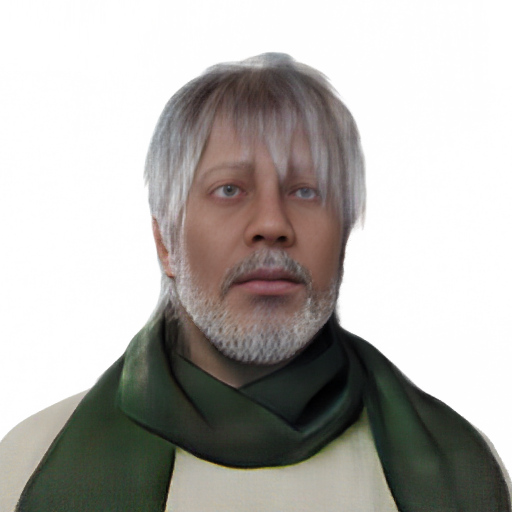}  \\        
    Base diffusion & Diffusion upsampler & Convolutional refiner \\
  \end{tabular}
  \vspace{-0.3em}
  \caption{Hierarchical generation progressively improves results.}
   \vspace{-2em}
  \label{fig:hierachical}
\end{figure}

\begin{figure*}[t]
  \centering
  \small
  \setlength\tabcolsep{1pt}
  \renewcommand{\arraystretch}{0.48}
  \begin{tabular}{ c@{}c@{\hspace{1mm}}c }    
    \includegraphics[width=0.375\linewidth]{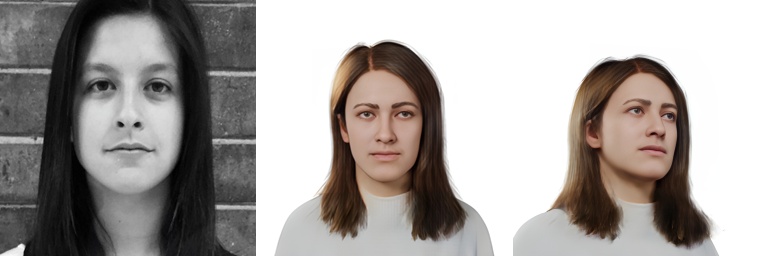} &  
    \includegraphics[width=0.375\linewidth]{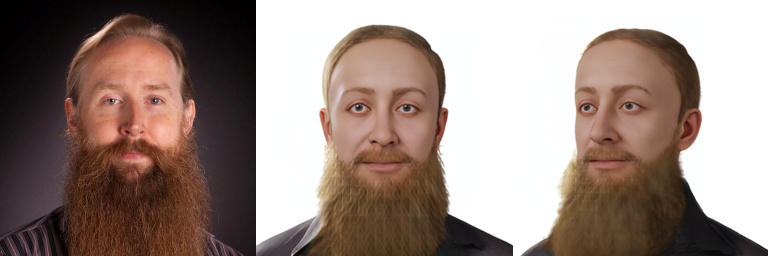} &
    \includegraphics[width=0.25\linewidth]{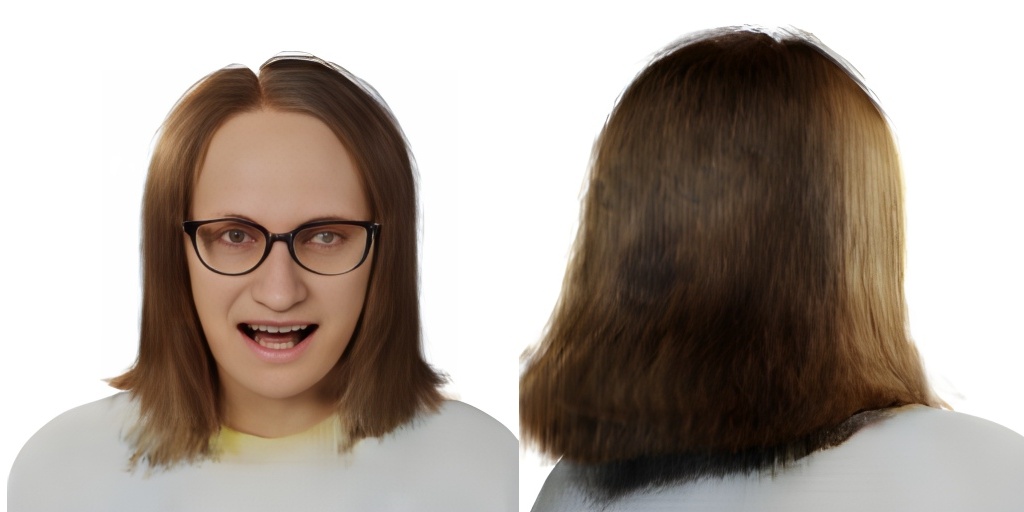}  \\    
    \put(292,-0.){\footnotesize ``\textit{Smiling young woman in glasses}"}   
    
    \\
    \includegraphics[width=0.375\linewidth]{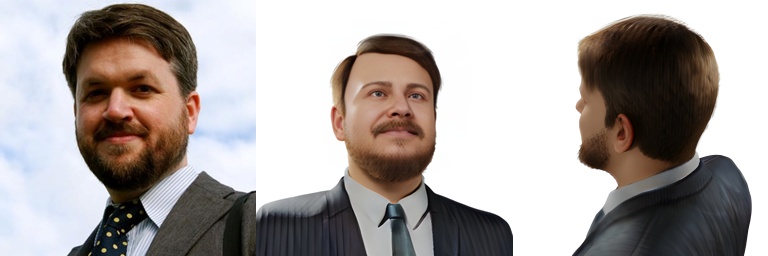} &  
    \includegraphics[width=0.375\linewidth]{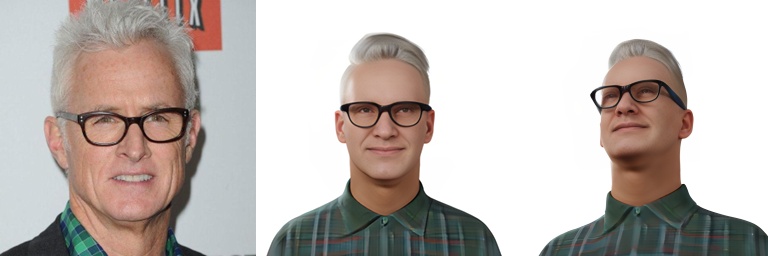} &
    \includegraphics[width=0.25\linewidth]{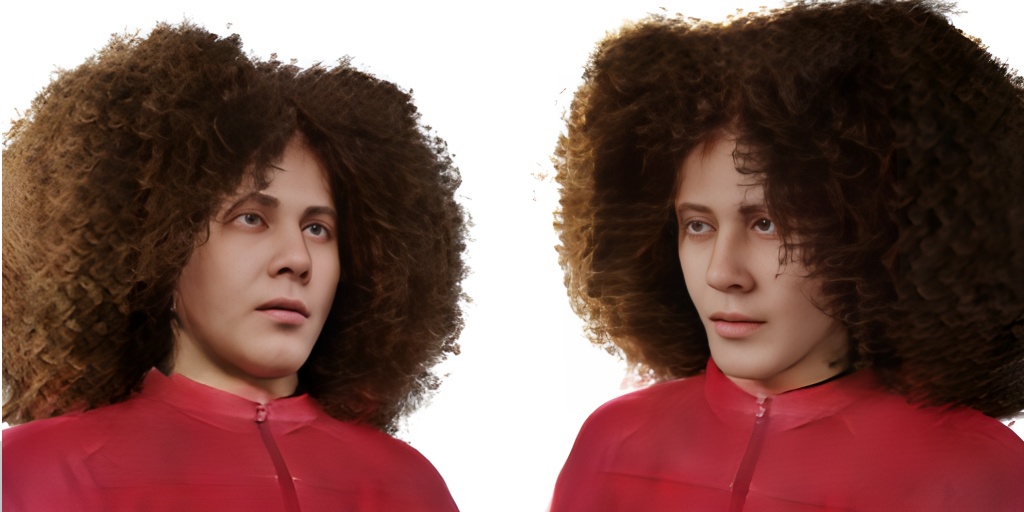}\\  
     \put(279,-0.0){\footnotesize ``\textit{A woman with afro hair in red wearing}"}
    \put(87,-8.5){\footnotesize (a)}
    \put(339,-8.5){\footnotesize (b)}    
  \end{tabular}

  \vspace{-0.5em}
  \caption{Results of (a) portrait inversion and (b) text-to-avatar generation.}
  \label{fig:portrait-inversion}
\end{figure*}

\subsection{Analysis of the Rodin model}

\noindent\textbf{Both 3D-aware convolution and latent conditioning are crucial for 3D synthesis.} To prove this, we conduct the ablation study as shown in Table~\ref{table:ablation}. We start from a baseline that uses a plain 2D CNN to process channel-wise concatenated tri-plane features following~\cite{chan2022efficient}. With latent conditioning, we achieve a lower FID. Feeding the network with roll-out tri-plane features significantly reduces the FID score because tri-planes are no longer improperly mingled. The proposed 3D-aware convolution further improves the synthesis quality, especially for thin structures like hair and cloth texture. More visual results regarding these ablations can be found in the Appendix.

\noindent\textbf{Hierarchical generation is critical for high-fidelity results.}
One significant benefit of this approach is that we can train different diffusion models dedicated to different scales in a supervised manner, as opposed to end-to-end synthesis with image loss.
This also enables patch-wise training without the need to render full images. Thus hierarchical training allows high-resolution avatar generation without suffering the prohibitive memory issue. Figure~\ref{fig:hierachical} shows the progressive quality improvement after the base diffusion, diffusion upsampler, and convolution refinement, respectively. It can be seen that the diffusion upsampler is critical, largely enhancing the synthesis quality, while convolution refinement further adds delicate details.

\begin{table}[]
  \centering
  \small
  \setlength\tabcolsep{3pt}
  \begin{tabular}{@{}ccccc@{} }
    \toprule
    Tri-plane level loss & Image-level loss & Cond. augment   & FID  $\downarrow$  \\
    \midrule
    \checkmark  &                 &            &  48.5 \\
    \checkmark  & \checkmark      &            &  38.6  \\
    \checkmark  & \checkmark      & \checkmark &  \textbf{26.1}   \\
    \bottomrule
  \end{tabular}
   \caption{Ablation study of the tri-plane upsampling strategy.}
   \label{table:ablation_sr_strategy}
\end{table}

\noindent\textbf{Diffusion upsampling training strategies.}
When training the tri-plane upsampler, we parameterize the model to predict the clean tri-plane ground truth at each diffusion reversion step. Meanwhile, conditioning augmentation is of great significance to let the model generalize to the coarse-level tri-plane generated from the base model. Besides, we observe enforcing image-level loss is beneficial to final perceptual quality. The effectiveness of these strategies are 
quantitatively justified in Table~\ref{table:ablation_sr_strategy}.

\subsection{Applications}
\label{sec:application}
\noindent\textbf{3D portrait from a single image.} We can hallucinate a 3D avatar from a single portrait by conditioning the base generator with the CLIP image embedding for that input image. Note that our goal is different from face/head reconstruction~\cite{deng2019accurate,ramon2021h3d}, but to conveniently produce a personalized digital avatar for users.  As shown in Figure~\ref{fig:portrait-inversion}(a), the generated avatars keep the main characteristics of the portrait, \eg, expression, hairstyle, glass wearing, etc., while being 360-degree renderable.

\begin{figure}[t]
  \centering
  \small
  \begin{overpic}
    [trim={0 0.0cm 0 0},clip,width=\columnwidth]{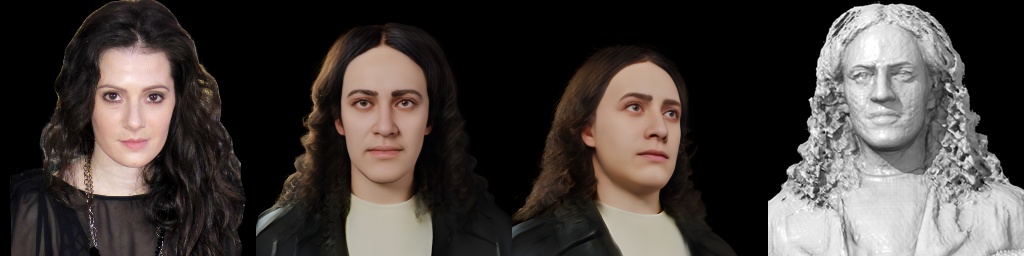}
    \put(10,-2.9){\footnotesize Input}
    \put(31.5,-2.9){\footnotesize Reconstructed avatar with visualized mesh}
  \end{overpic}  
 \vspace{0.03em}
 
  \begin{overpic}
    [trim={0 0.0cm 0 0},clip,width=\columnwidth]{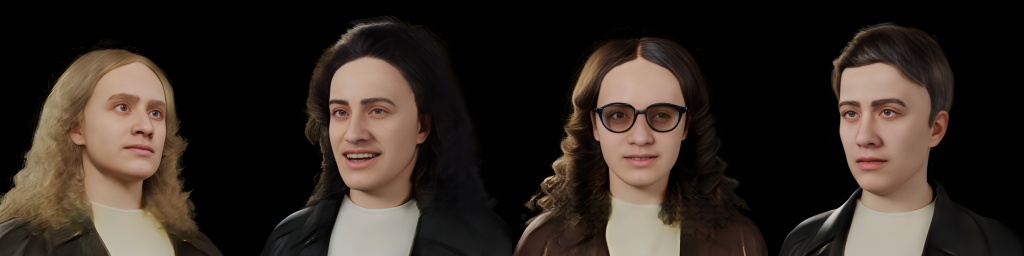}
    \put(4,-2.9){\footnotesize ``\textit{Blond hair}''}
    \put(28.5,-2.9){\footnotesize ``\textit{Smiling}''}
    \put(50,-2.9){\footnotesize ``\textit{With sunglasses}''}
    \put(80,-2.9){\footnotesize ``\textit{Short hair}''}
  \end{overpic}  
  \\
  \vspace{0.6em}
  \caption{Results of text-guided avatar manipulation.}
  \vspace{-2mm}
  \label{fig:edit}
\end{figure}

\noindent\textbf{Text-to-avatar generation.} Another natural way to customize avatars is to use language guidance. To do this, we train a text-conditioned diffusion model to generate the CLIP image embedding used to semantically control the avatar generation. We use a subset of the LAION-400M dataset~\cite{schuhmann2021laion} containing  portrait-text pairs to train this model. As shown in Figure~\ref{fig:portrait-inversion}(b), one can finely customize the avatars using detailed text descriptions.

\noindent\textbf{Text-based avatar customization.}
We can also semantically edit generated avatars using text prompts. For a generated avatar with the CLIP image embedding $\bm{z}_i$, we can obtain a direction $\bm{\delta}$ in the CLIP's text embedding based on prompt engineering~\cite{patashnik2021styleclip}. We assume colinearity between the CLIP's image and text embedding, thus we obtain the manipulated embedding as $\bm{z}_i+\bm{\delta}$, which is used to condition the generative process. As shown in Figure~\ref{fig:edit}, we can achieve a wide variety of disentangled and meaningful control faithful to the text prompt.

\section{Conclusion}
\label{sec:conclusion}

From experiments, we observe that the Rodin model is a powerful generative model for 3D avatars. This model also allows users to customize avatars from a portrait or text, thus significantly lowering the barrier of personalized avatar creation. While this paper only focuses on avatars, the main ideas behind the Rodin model are applicable to the diffusion model for general 3D scenes. Indeed, the prohibitive computational cost has been a 
challenge for 3D content creation. An efficient 2D architecture for performing coherent and 3D-aware diffusion in 3D is an important step toward tackling the challenge. For future work, it would be fruitful to improve the sampling speed of the 3D diffusion model and study jointly leveraging the ample 2D data to mitigate the 3D data bottleneck. 

{\small
\bibliographystyle{ieee_fullname}
\bibliography{egbib}
}

\section*{Appendix}
\appendix 
\section{Background of Diffusion Models}
Diffusion models produce data by reversing a gradual noising process.
The forward noising process is a Markov chain that corrupts the data by gradually adding random noises for steps $t=1, \cdots, T$. Each step in the forward process is a Gaussian transition $q(\xb_t|\xb_{t-1}) :=\mathcal{N}(\sqrt{1-\beta_t}\xb_{t-1}, \beta_t\bm{I})$, where $\{\beta_t\}^T_{t=0}$ are usually pre-defined variance schedule. Furthermore, the noisy latent variable $\xb_t$ can be derived from $\xb_0$ directly as:
\begin{align}
    \xb_t = \sqrt{\alpha_t}\xb_0 + \sqrt{1 - \alpha_t}\zb , \ \ \ \zb \sim \mathcal{N} (\bm{0},\bm{I}),
\end{align}
where $\alpha_t := \prod_{s=1}^{t} {(1-\beta_s)}$. When $T$ is large enough, $\alpha_T$ gets closer to $0$ and the last latent variable $\xb_T$ is nearly an isotropic Gaussian distribution.

To sample data from the given distribution, we can reverse the noising process by learning a denoising model $\bm{\epsilon}_{\theta}(\bm{x}_{t},t)$.  The denoising model $\bm{\epsilon}_{\theta}(\bm{x}_{t},t)$ starts from the Gaussian noise $\bm{x}_T$ and iteratively reduces the noise for $t=T-1, \cdots, 0$. Specifically, it takes the noisy latent variable $\bm{x}_t$ at each timestep $t$ and predicts the corresponding noise~$\bm{\epsilon}$ with a minimal mean square error:
 \begin{align}
     \min_{\theta}\mathbb{E}_{\xb_0 \sim p(\xb_0), \zb \sim \mathcal{N}(\bm{0},\bm{I}), t} ||\epsilonb_\theta(\xb_t, t) - \zb||^2_2.
\end{align}
With the learned denoising model, the data can be sampled with the following reverse diffusion process:
\begin{align}
    \label{eq:reverse_ddpm}
    \xb_{t-1} = \frac{1}{\sqrt{1-\beta_t}}\left(\xb_t - \frac{\beta_t}{\sqrt{1-\alpha_t}} \epsilonb_\theta(\xb_t, t)\right) + \sigma_t\zb,
\end{align}
where $\zb \sim \mathcal{N} (\bm{0},\bm{I})$ is a randomly sampled noise, and $\sigma_t$ is the variance of the added noise.

\section{Implementation Details}
\subsection{Architectural Design and Training Details}
\begin{figure}
    \centering
    \includegraphics[width=0.99\linewidth]{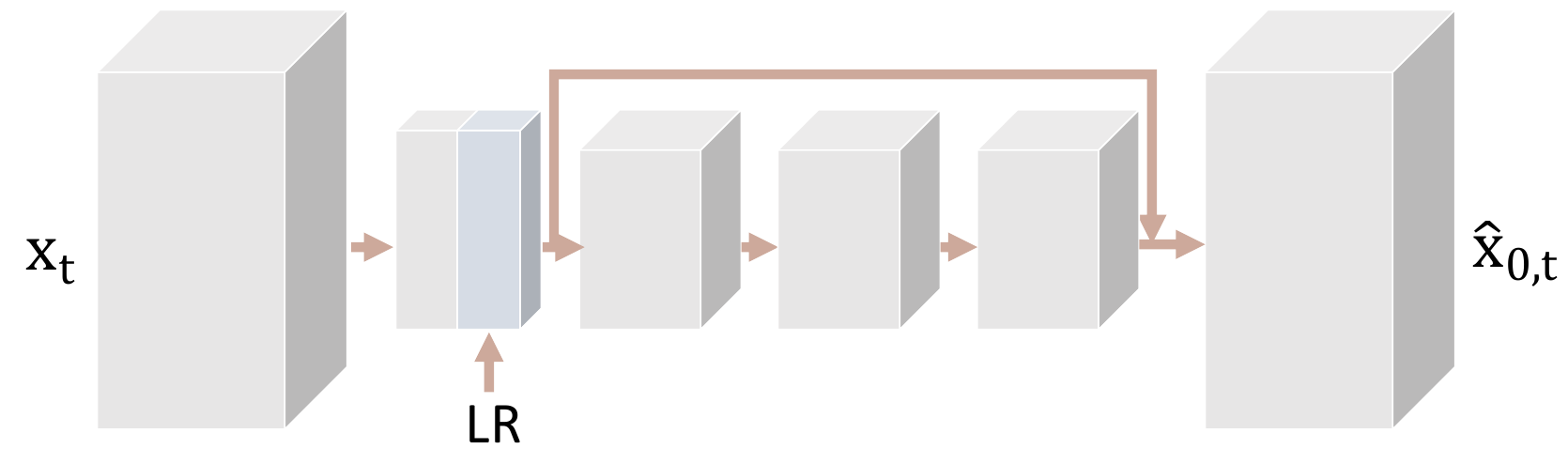}
    \caption{Architecture of the upsample diffusion model.}
    \label{fig:arch_upsample}
\end{figure}

Our base diffusion model adopts the U-Net architecture from~\cite{dhariwal2021diffusion} with a channel number of 192, while we make several modifications including tri-plane roll-out and 3D-aware convolution, as discussed in Section \ref{sec:latent conditioned}. To orchestrate the tri-plane generation and enable semantic editing, we also introduce a condition encoder,  a fixed CLIP ViT-B/32 image encoder, to map a reference image to a semantic latent vector.  The upsample diffusion model is a U-Net-like model but we  apply only one upsample layer that directly upscales the feature maps from $64$ to $256$ for efficiency, as shown in Figure~\ref{fig:arch_upsample}. The tri-plane roll-out and 3D-aware convolution are utilized in each residual block. When training the upsample model, we apply condition augmentation on the tri-planes to reduce the domain gap as described in Section \ref{sec:tri-plane upsampler}. Specifically, we degrade the ground-truth $256\times256$ tri-planes with a random combination of downscale, Gaussian blur, and Gaussian noise. 

We utilize AdamW optimizer~\cite{adamw} with a batch size of 48 and a learning rate of $5e\textnormal{-}5$ for the base diffusion model, and with a batch size of 16 and a learning rate of $5e\textnormal{-}5$ for the upsample diffusion model. We also apply the exponential moving average (EMA) with a rate of $0.9999$ during training. We set the diffusion steps as 1,000 for the base model, and 100 for the upsample model, with a linear noise schedule. During inference we sample $100$ diffusion steps for both the base model and the upsample model. All the experiments are performed on NVIDIA Tesla 32G-V100 GPUs.

\subsection{Tri-plane Fitting}
Our framework learns the 3D avatar generation from explicit 3D representations obtained from fitting multi-view images.
However, a multi-view consistent, diverse, high-quality and large-scale dataset of face images is difficult and expensive to collect. Images collected from the Web have no guarantee of multi-view consistency and suffer privacy and copyright risks. Regarding this, we turn to synthetic techniques that can randomly render novel 3D portraits by randomly combining assets manually created by artists. We leverage the Blender synthetic pipeline~\cite{wood2021fake} that generates human faces along with random sampling from a large collection of hair, clothing, expression and accessory. Hence, we create 100K synthetic individuals independently and for each render 300 multi-view images with a resolution of $256 \times 256$.

\begin{figure}
    \centering
    \includegraphics[width=0.99\linewidth]{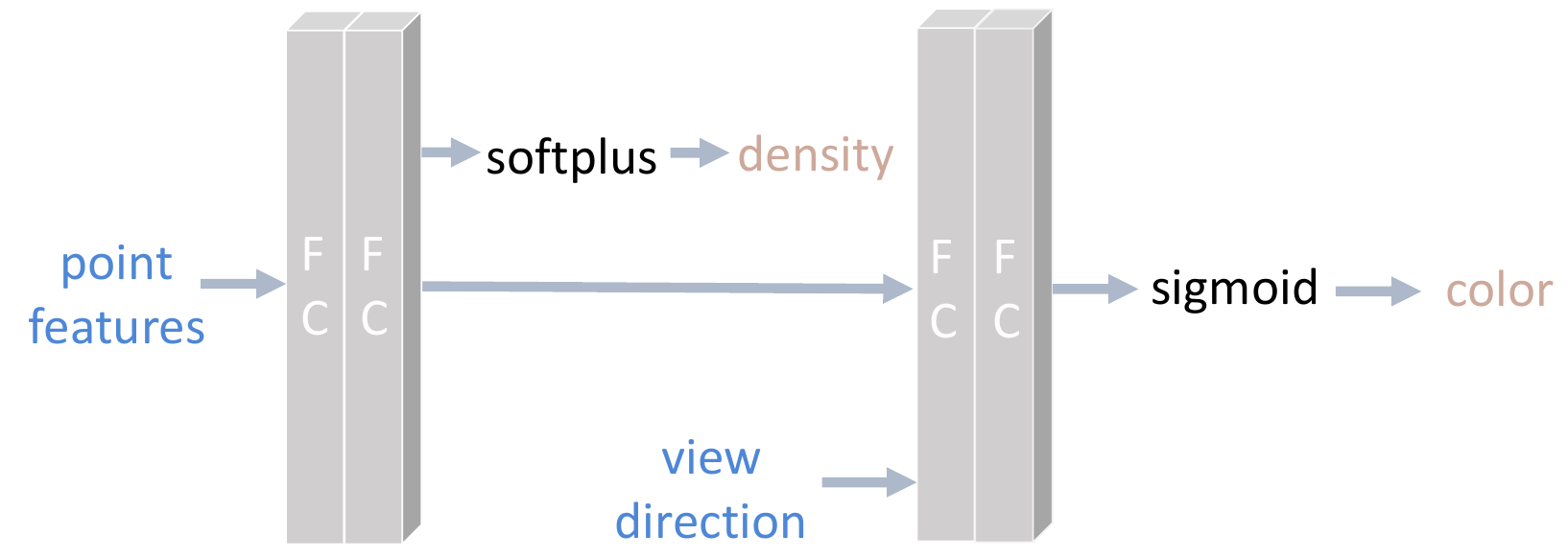}
    \caption{Architecture of the MLP decoder.}
    \label{fig:arch_renderer}
\end{figure}

For tri-plane fitting, we learn $256\times 256\times 32\times 3$ spatial features for each person along with a lightweight MLP decoder consisting of 4 fully connected layers as shown in Figure~\ref{fig:arch_renderer}.  We randomly initialize the tri-plane feature and MLP weights.  During fitting, we apply random rescaling (downsample to a resolution in $[64,256]$ followed by an upsampling to $256$) to ensure that the MLP decoder is robust to multi-resolution tri-plane features. To enable scalable and efficient fitting, we first optimize the shared 4-layer MLP decoder  when fitting the first 1,000 subjects, and this decoder is fixed when fitting the following subjects. Thus different subjects are fitted separately in distributed servers.

For multi-view images $\{\bm{x}\}_{N_v}$ for the given subject, where $\bm{x}\in \mathbb{R}^{H_0\times W_0 \times 3}$, we minimize the mean squared error $\mathcal{L}_{\textnormal{MSE}}$
between the rendered image via volumetric rendering, \ie, $\hat{\bm{x}}=\mathcal{R}(\bm{c},\sigma)$ and the corresponding ground truth image. Moreover, we introduce additional regularizers to
improve the fitting quality. To be specific, we manage to reduce the ``floating'' artifact by enforcing the sparsity loss $\mathcal{L}_{\textnormal{sparse}}$ which penalizes the
$\ell_1$ magnitude of the predicted density, the smoothness loss $\mathcal{L}_{\textnormal{smooth}}$~\cite{chan2022efficient} that encourages a smooth density field, as well as the distortion loss
$\mathcal{L}_{\textnormal{dist}}$~\cite{barron2022mip}   that encourages compact rays with localized weight distribution.

\subsection{Text-based Avatar Customization}
As shown in Section \ref{sec:application}, the Rodin model can edit generated avatars with text prompts. For a generated avatar with a conditioned latent $\bm{z}_i$, we can obtain an editing direction $\bm{\delta} = E_T^{clip}(T_{tgt})- E_T^{cilp}(T_{src})$ in the text embedding space of CLIP based on prompt engineering. For instance, we can choose the source text $T_{src}$ from some general descriptions such as ``a photo of a person" and ``a portrait of a person", and use the target text $T_{tgt}$ such as ``a photo of a person with blond hair" and ``a photo of a smiling person ". As we assume colinearity between the CLIP's image and text embedding,  we can obtain the manipulated embedding as $\bm{z}_i+\bm{\delta}$, which is used to generate edited avatars.

\subsection{Latent Diffusion for Unconditional Sampling}
As discussed in Section \ref{sec:latent conditioned}, our base diffusion model supports both unconditional generation and conditional generation. To account for full diversity during unconditional sampling, we additionally train a diffusion model to model the distribution of the latent $\bm{z}$. The latent diffusion adopts a 20-layer MLPs network~\cite{preechakul2021diffusion} with the hidden channel of 2048 that iteratively predicts the latent code $\bm{z}\in \mathbb{R}^{512}$ from random Gaussian noise. We set the diffusion steps as 1,000 with a linear noise schedule. We utilize AdamW optimizer with a batch size of 96 and a learning rate of $4e-5$, and also apply exponential moving average (EMA) with a rate of 0.9999 during training.

\begin{figure}[t]
    \centering
     \vspace{-5mm}
 \includegraphics[width=\linewidth]{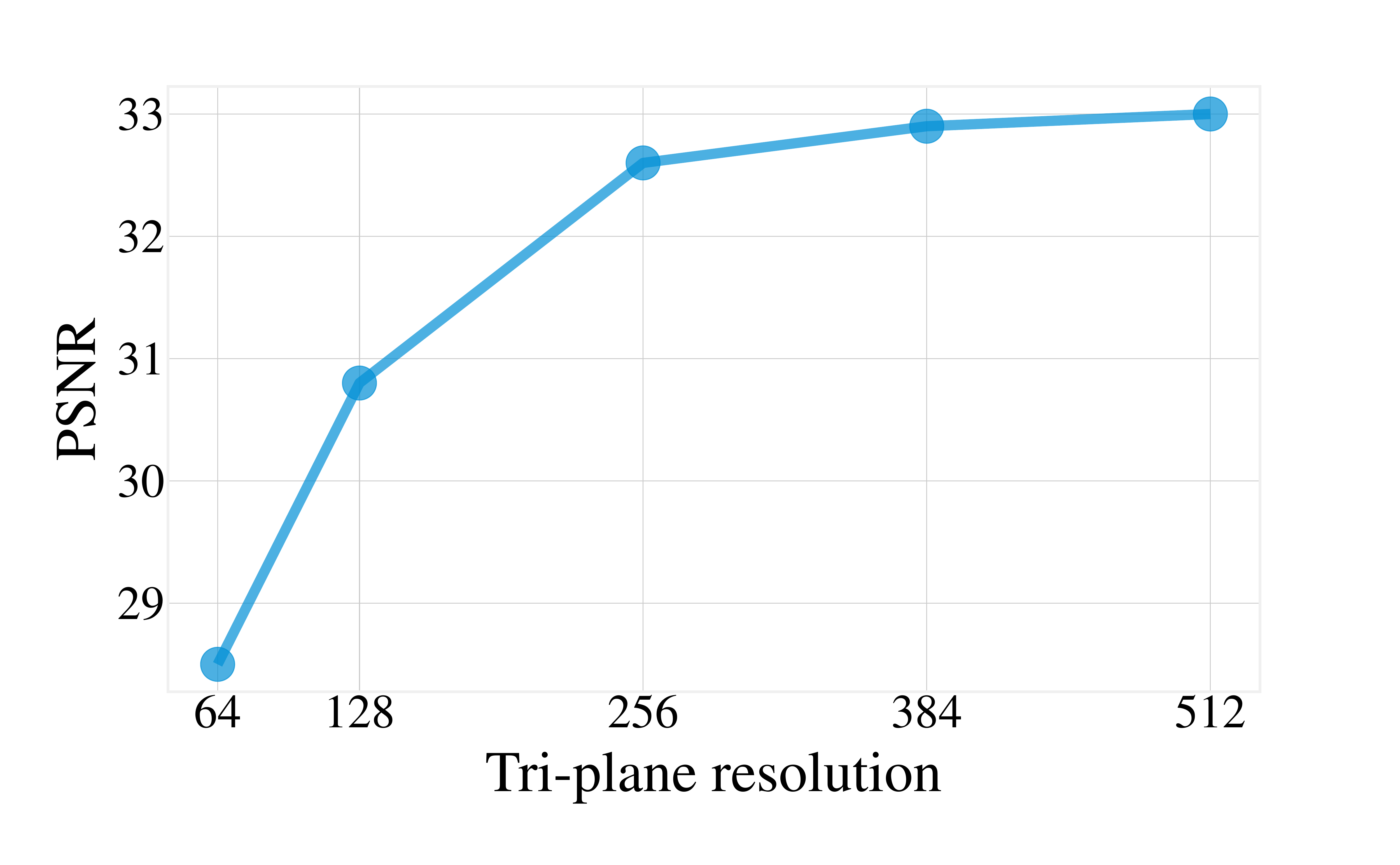}
    \vspace{-5mm}
    \caption{Effect of tri-plane resolution for tri-plane fitting.}
    \label{fig:psnr-triplane}
\end{figure}

\begin{figure}[t]
    \centering
     \vspace{-5mm}
\includegraphics[width=\linewidth]{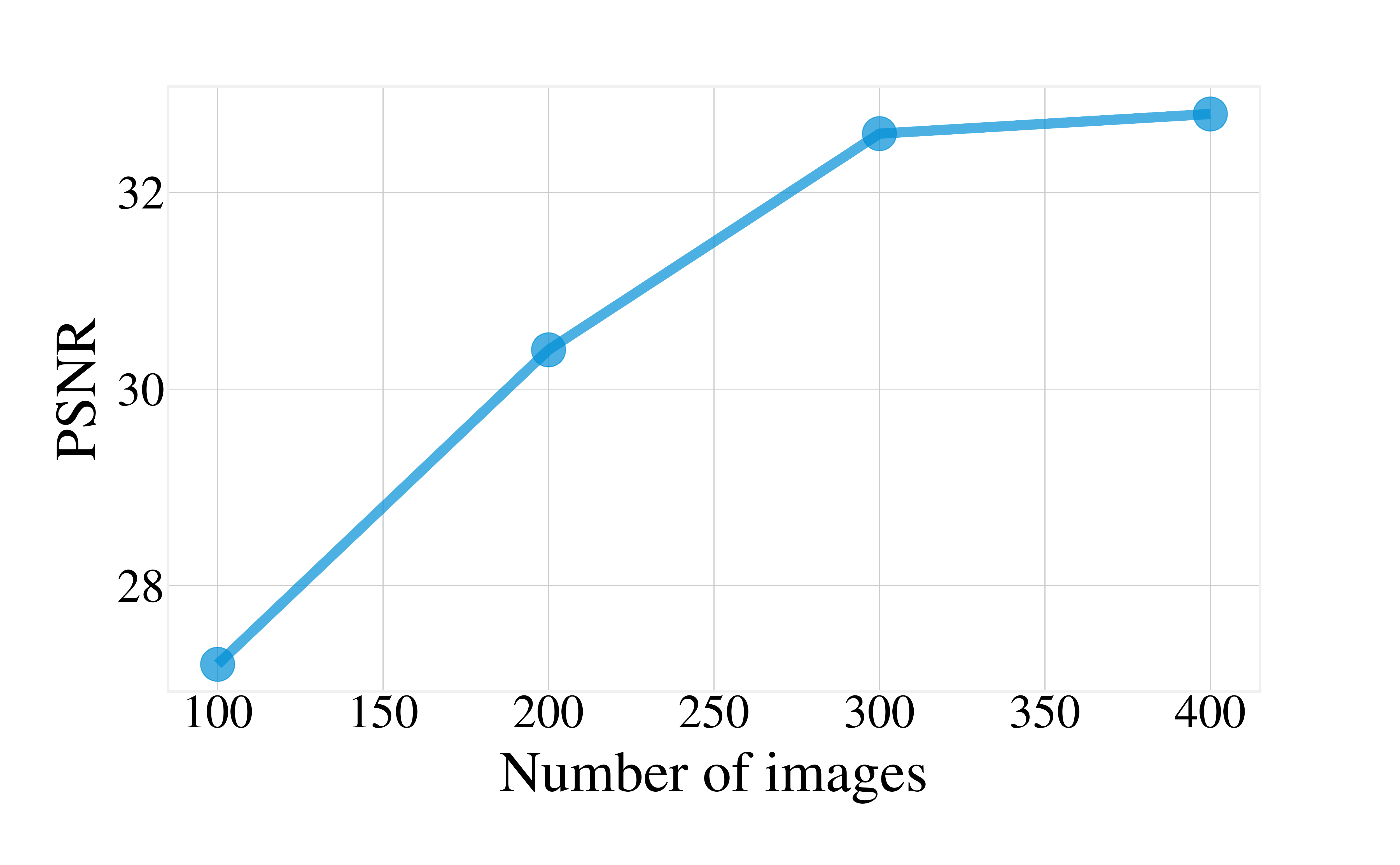}
 \vspace{-5mm}
    \caption{Effect of  image numbers for tri-plane fitting.}
    \label{fig:psnr-image}
\end{figure}

\subsection{Text-to-avatar Generation}
As shown in Section \ref{sec:application},  we perform text-to-avatar generation by training a text-conditioned diffusion model that generates an image embedding  from a text embedding in the CLIP space. We adopt the network architecture from~\cite{ramesh2022hierarchical} and train it on a subset of the LAION-400M dataset, containing 100K portrait-text pairs. We set the diffusion steps as 1,000 with a linear noise schedule. We utilize AdamW optimizer with a batch size of 96 and a learning rate of $4e-5$, and also apply exponential moving average (EMA) with a rate of 0.9999 during training.

\section{Additional Ablation Study and Analysis}
\subsection{Tri-plane Settings}

\noindent\textbf{Choices of Tri-plane resolution.} To analyze the impact of tri-plane resolution, we experiment with different tri-planes from a set of $\{64, 128, 256, 384, 512\}$ to fit $1024 \times 1024$ images and show the results in Figure~\ref{fig:psnr-triplane}. Overall, the fitting quality increases with the tri-plane resolution. Empirically, we find that the $256\times256$ tri-plane is strong enough to represent a subject. Considering the memory cost, we thus choose to utilize $256\times256$ tri-planes in our experiments.

\begin{figure}[t]
    \centering
    \includegraphics[width=0.96\linewidth]{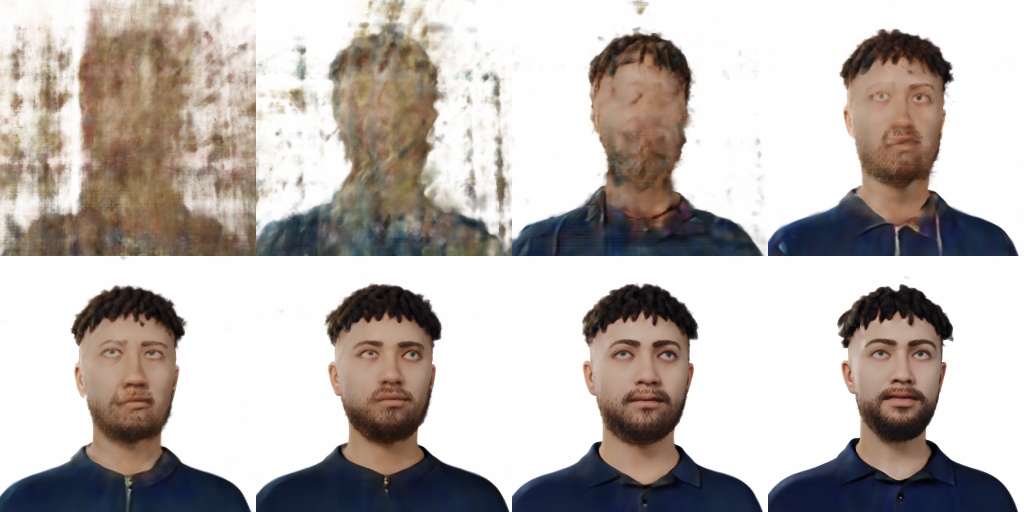}
    \caption{Visualization of intermediate generation results of different time steps.}
    \label{fig:steps}
\end{figure}

\begin{figure}[t]
  \centering
  \small
  \setlength\tabcolsep{1pt}
  \renewcommand{\arraystretch}{0.5}
  \begin{tabular}{@{}c@{}}    
    \includegraphics[width=0.92\linewidth]{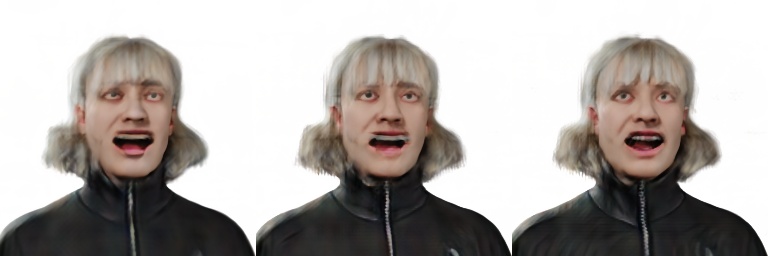} \\
    \begin{overpic}
    [width=0.46\textwidth]{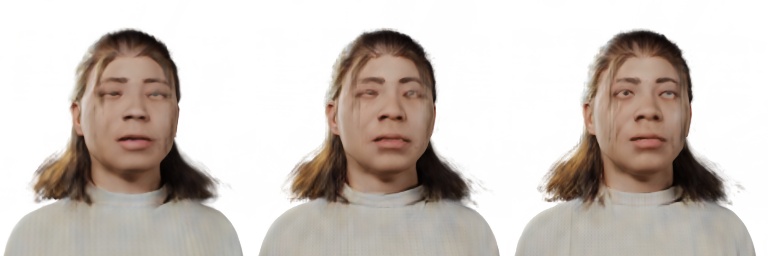}
    \put(14,-4){\footnotesize Base}
    \put(40,-4){\footnotesize  $+$ Roll-out}
    \put(70,-4){\footnotesize   $+$ 3D-aware conv}
     \end{overpic} 
     \vspace{0.9em}
  \end{tabular}
  \caption{Both tri-plane roll-out and 3D-aware convolution are crucial for high-fidelity results.}
  \label{fig:ablation-arch}
\end{figure}

\begin{figure*}[h]
  \centering
  \small
  \setlength\tabcolsep{1pt}
  \renewcommand{\arraystretch}{0.5}

    \begin{overpic}
    [width=0.95\textwidth]{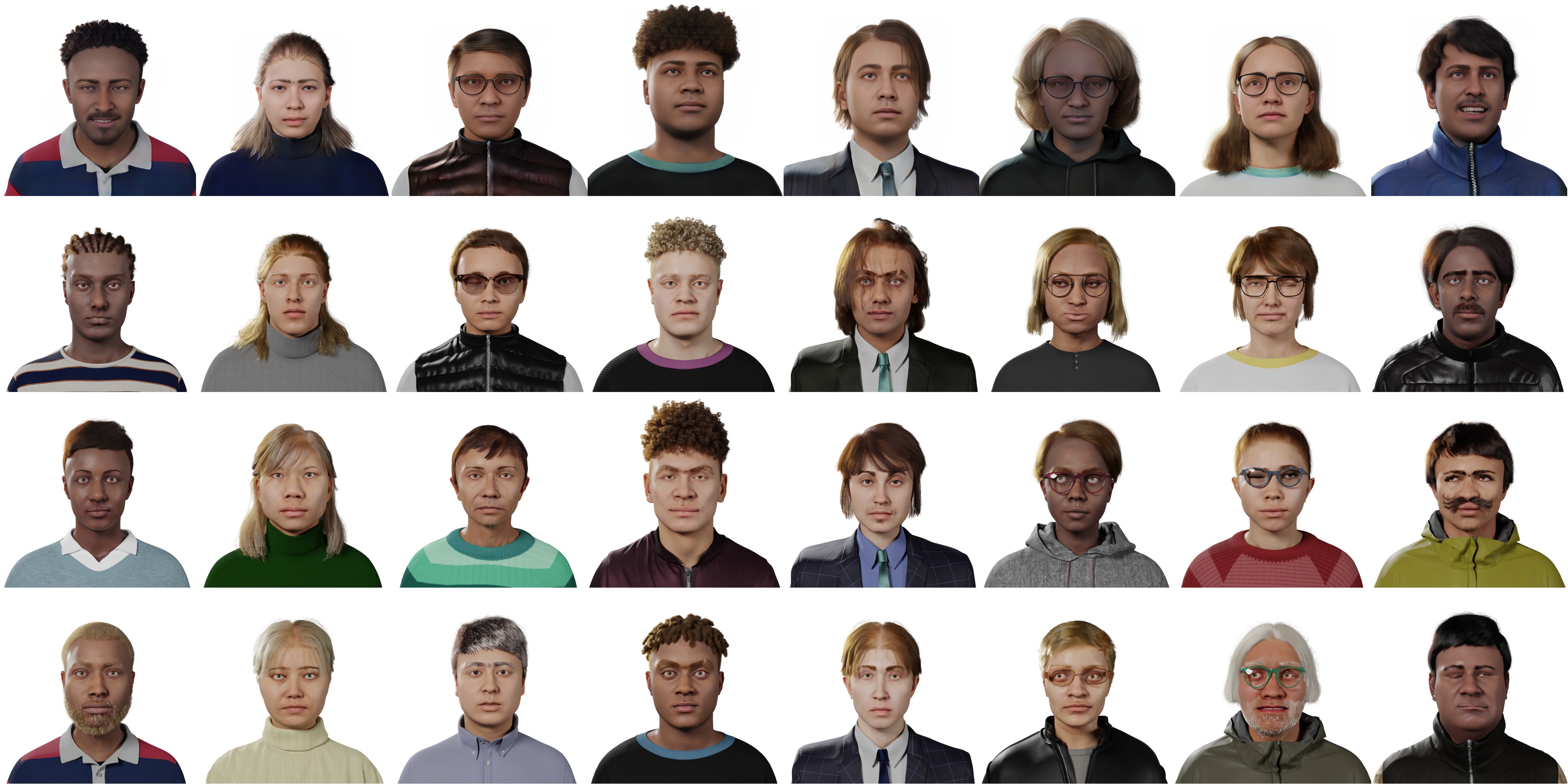}
    \put(-2,39){\footnotesize \rotatebox{90}{Generated}}
    \put(-2,10){\footnotesize  \rotatebox{90}{Top-3 nearest neighbors}}
     \end{overpic} 
     \vspace{0.2em}
  \caption{Nearest neighbors in the training data according to CLIP feature similarity.}
  \label{fig:nn}
\end{figure*}

\noindent\textbf{Number of images for fitting.} We also explore how many images are needed to achieve a high-quality fitting. As shown in Figure~\ref{fig:psnr-image}, the fitting quality get almost saturated when using 300 different views for the neural tri-plane reconstruction.

\subsection{Visualization of Different Diffusion Steps}
Diffusion models generate samples by gradually removing noises for $t\in [T, 0]$, and analyzing these intermediate results would reach an in-depth understanding of the generation process. We thus demonstrate the generated results over the reverse process in Figure~\ref{fig:steps}, where we render the predicted tri-plane of our base diffusion, $\hat{x_0}$, at each time step $t$. Notwithstanding that our diffusion is performed in tri-plane feature space, the reverse process is similar to that in image space, where the coarse structure appears first and fine details appear in the last iterative steps.

\subsection{Effect of 3D-aware Convolution}
By rolling out tri-plane feature maps and applying 3D-aware convolution, the Rodin model performs 3D-aware diffusion using an efficient 2D architecture.  As analyzed in Section \ref{sec:latent conditioned}, tri-plane roll-out and 3D-aware convolution are crucial for high-fidelity results, especially for thin structures such as hair strands and clothing details, by enhancing cross-plane communication. To validate the impact of these designs in high-quality tri-plane, we modify the upsample diffusion model with different configurations and remove the convolution refinement with the base diffusion fixed. Figure~\ref{fig:ablation-arch} demonstrates with rollout and 3D-aware convolution, the full model shows a clear improvement compared to the base model.

\begin{table}[tt]
  \centering
  \small
  \setlength\tabcolsep{3pt}
  \begin{tabular}{@{}l@{\hspace{5mm}}c@{\hspace{5mm}}c@{\hspace{5mm}}c@{\hspace{6mm}}c@{\hspace{6mm}}c@{} }
    \toprule
  Scale & w/o CFG & 1.2  &  1.5  & 3.0  & 6.0\\
    \midrule
    PSNR  &  24.06   &   \textbf{24.21}       &  24.07    &  24.05     & 24.15  \\
    SSIM  &    \textbf{0.795}  &     0.794     &   0.792   &   0.782    & 0.775   \\
    LPIPS  &    0.128  &     \textbf{0.121}       &   0.133   &    0.141   &  0.146 \\
    \bottomrule
  \end{tabular}
   \caption{Quantitative results of conditional avatar reconstruction.}
   \label{table:cfg}
\end{table}

\subsection{Nearest Neighbors Analysis}
The Rodin model enables a hassle-free creation experience of an unlimited number of avatars from scratch, each avatar being distinct. Figure~\ref{fig:nn} shows the nearest neighbors of some generated samples in the main paper, which indicates that the model does not simply memorize the training data.

\begin{figure*}[h]
  \centering
  \small
  \setlength\tabcolsep{1pt}
  \begin{tabular}{@{}c@{}}
    \includegraphics[width=0.99\linewidth]{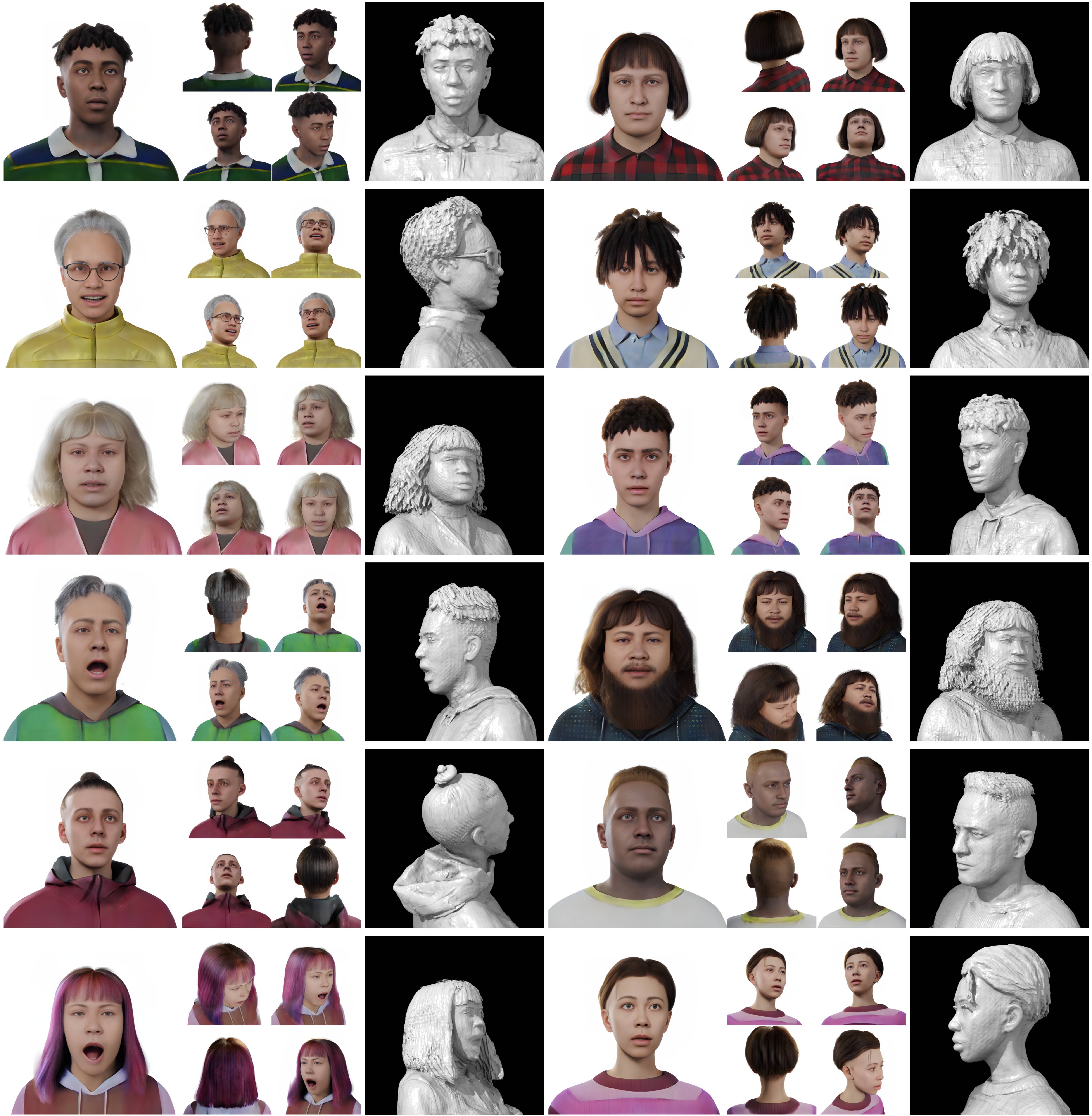} \\
    
  \end{tabular}
  \vspace{-0.5em}
  \caption{Unconditional generation samples by our Rodin model. We visualize the mesh extracted from the generated density field.}
  \label{fig:main-results-more}
\end{figure*}

\begin{figure*}[h]
  \centering
  \small
  \setlength\tabcolsep{1pt}
  \begin{tabular}{@{}c@{}}
    \includegraphics[width=0.99\linewidth]{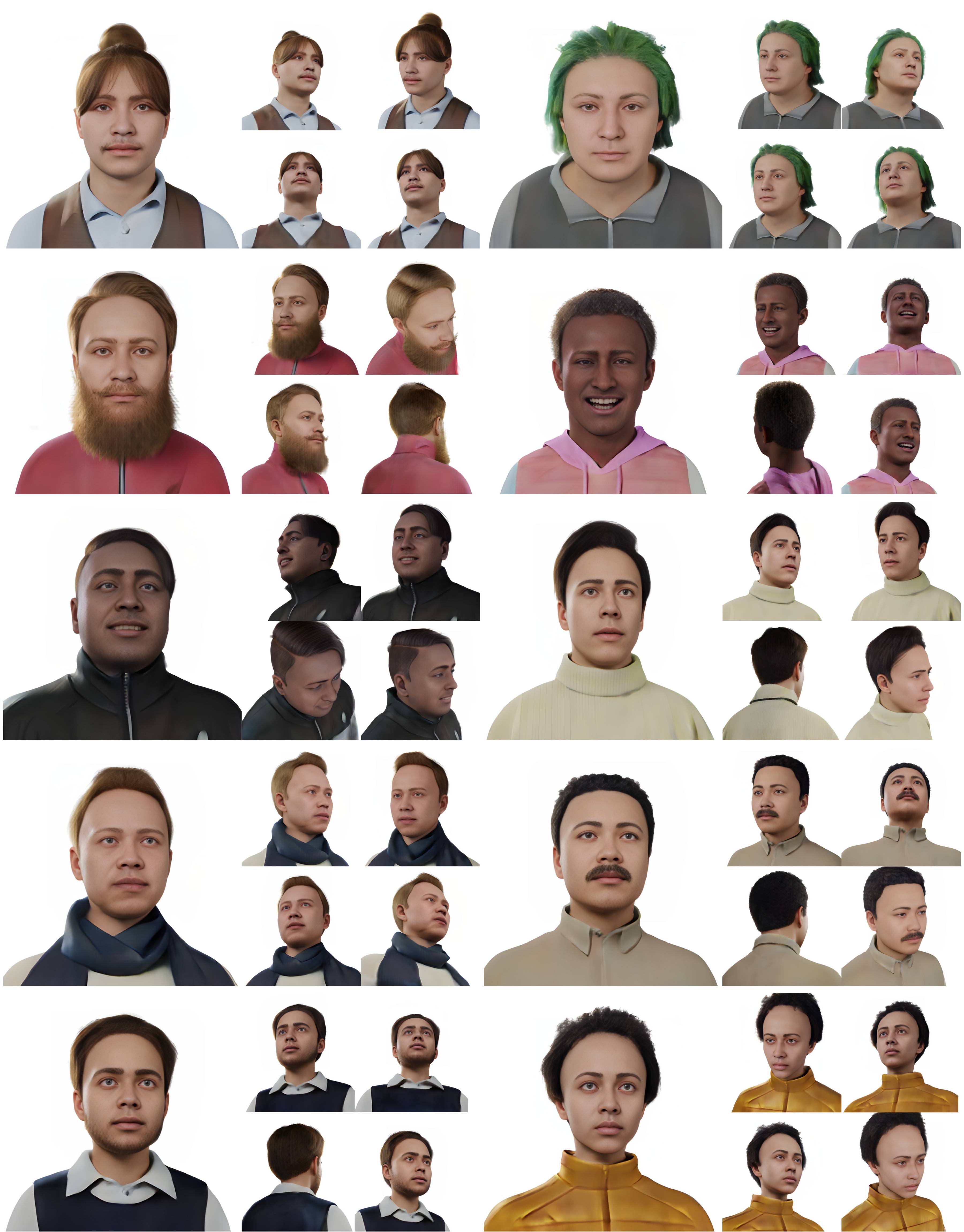}    \\     
  \end{tabular}
  \vspace{-0.5em}
  \caption{Unconditional generation samples by our Rodin model.}
  \label{fig:main-results-more2}
\end{figure*}

\begin{figure*}[h]
    \centering
    \includegraphics[width=0.96\linewidth]{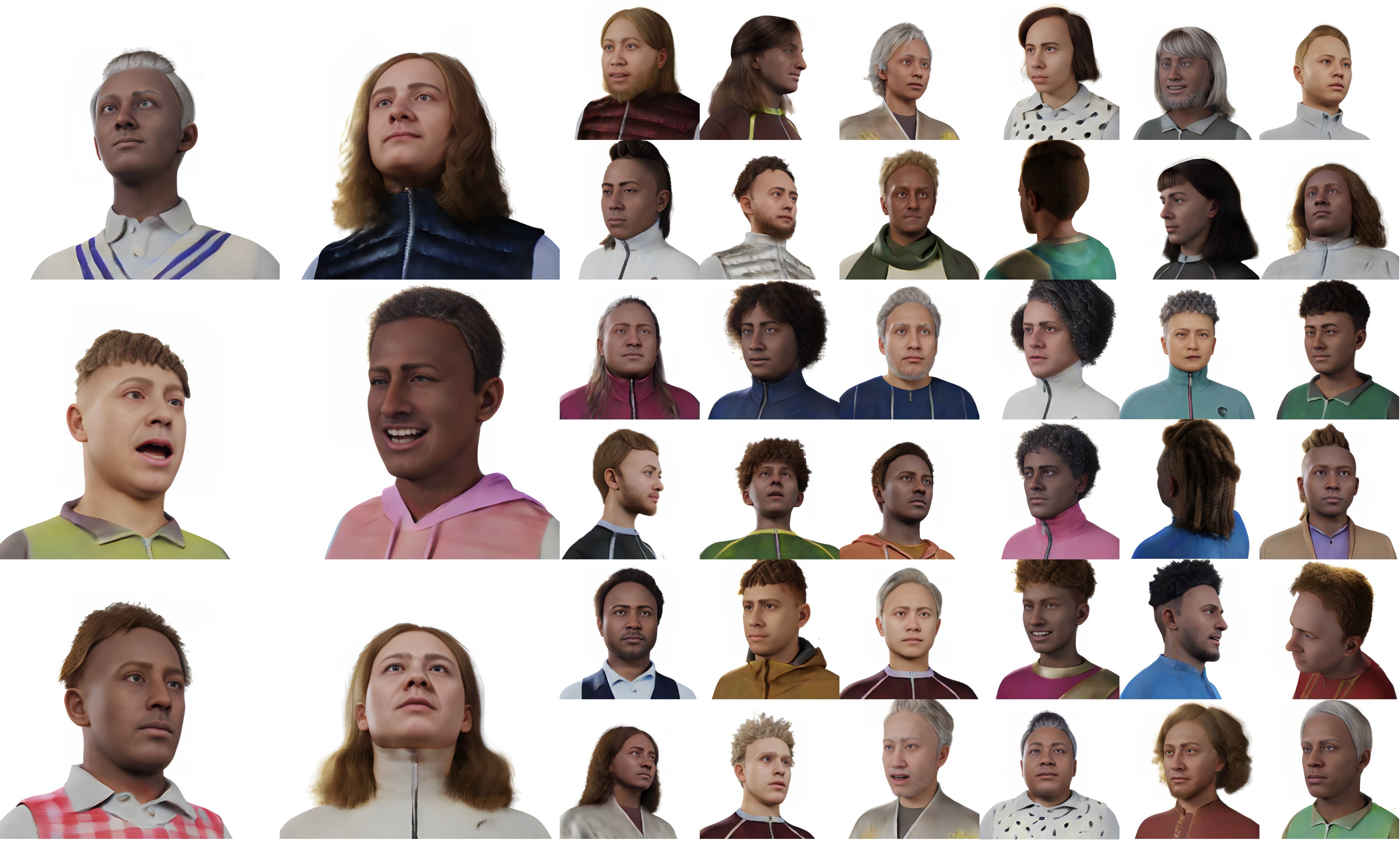}
    \caption{Uncurated  generation results by our Rodin model.}
    \label{fig:random}
\end{figure*}

\begin{figure*}[h]
  \centering
  \small
  \setlength\tabcolsep{1pt}
  \begin{tabular}{@{}c@{}} 

    \includegraphics[width=0.99\linewidth]{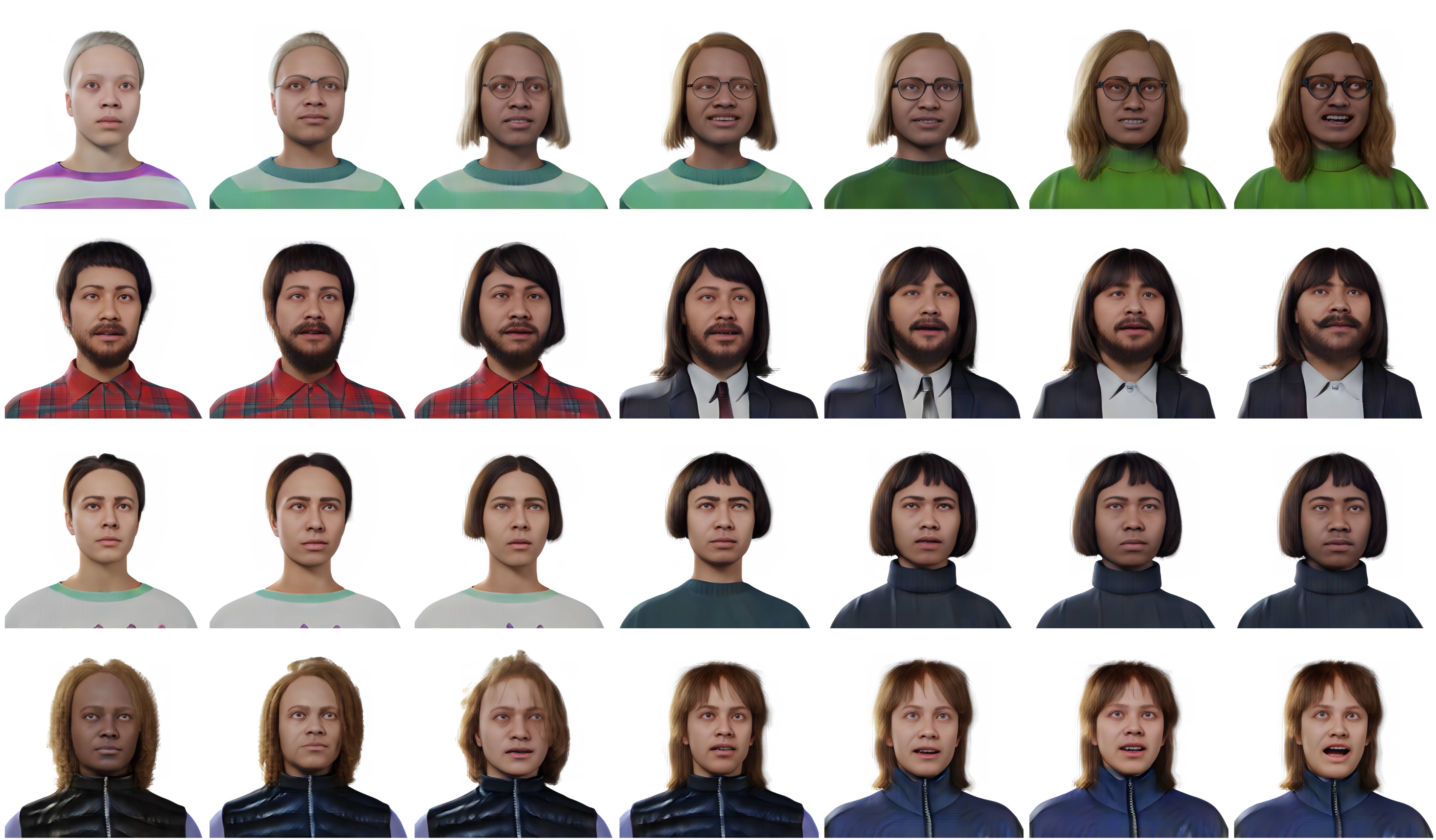}\\
  \end{tabular}
  \vspace{-0.5em}
  \caption{Latent interpolation results for generated avatars.}
  \label{fig:interpolation-more}
\end{figure*}

\begin{figure*}[h]
  \centering
  \small
  \setlength\tabcolsep{1pt}
  \renewcommand{\arraystretch}{0.5}
  \begin{tabular}{@{}c@{}}    
    \includegraphics[width=0.99\linewidth]{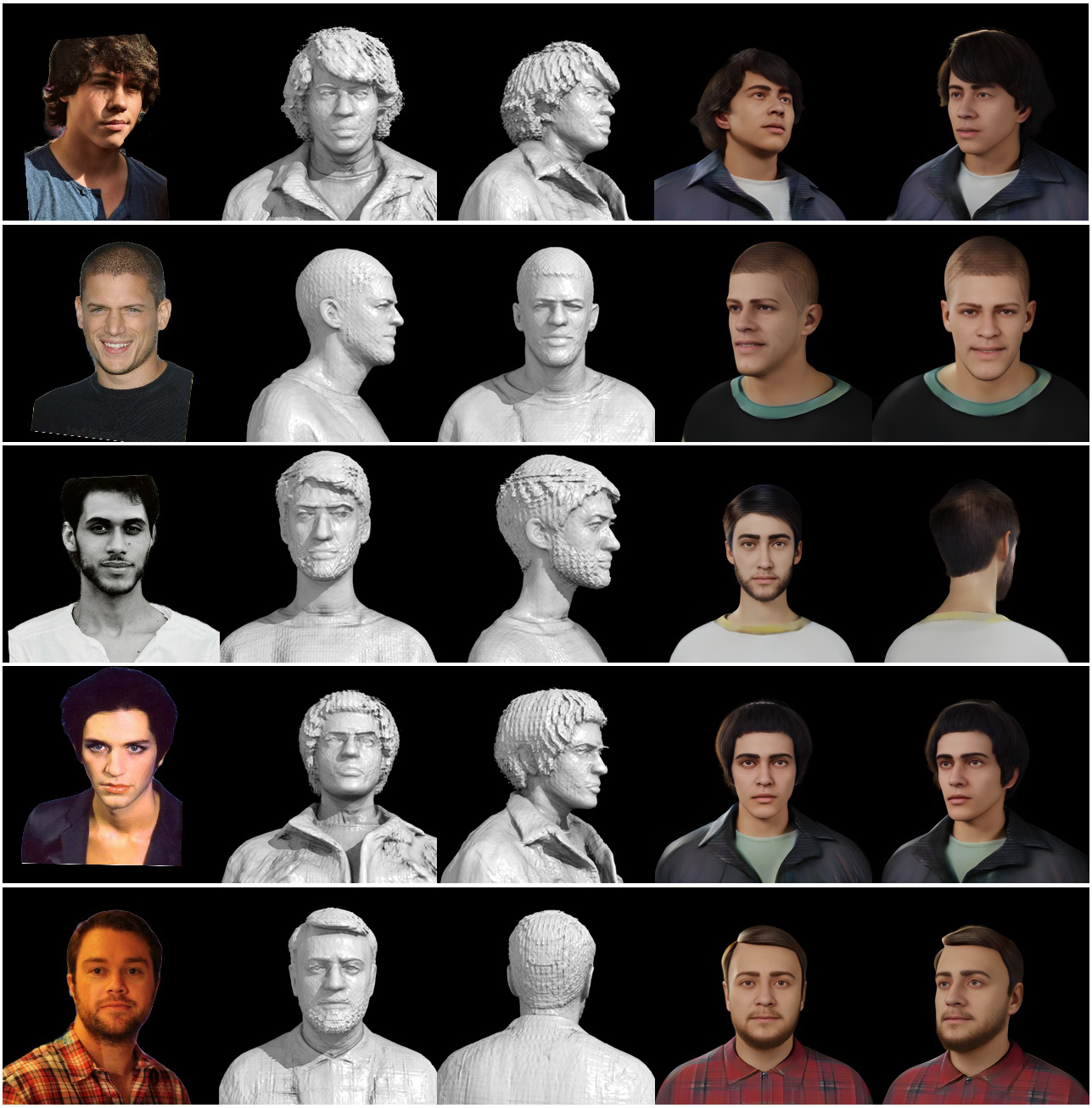} \\  
        
  \end{tabular}
  \vspace{-0.5em}
  \caption{Additional results of portrait inversion.}
  \label{fig:portrait-inversion-more}
\end{figure*}

\subsection{Conditional Avatar Generation} 
\noindent\textbf{Quantitative metrics.}
On top of unconditional generation, we can also hallucinate a 3D avatar from a single reference image by conditioning the base generator with the CLIP image embedding for that input image. We evaluate the conditional generation on 1K test data where each subject contains 300 images from different views. Table~\ref{table:cfg} reports the metrics between reconstructed images and ground-truth synthetic images.

\noindent\textbf{Classifier-free guidance.}
Our model supports classifier-free guidance (CFG) sampling when inference, which is a technique typically used to boost the sampling quality in conditional generation. Table~\ref{table:cfg} evaluates generation quality with different scales of classifier-free guidance in terms of PSNR, SSIM and LPIPS.

\section{Additional Visual Results}
Figure~\ref{fig:main-results-more} and Figure~\ref{fig:main-results-more2} show more random samples generated by the Rodin model, showing the capability to synthesize high-quality 3D renderings with impressive details.  To reflect the geometry, we also extract the mesh from the generated density field using marching cubes, which demonstrates high-fidelity geometry. Figure~\ref{fig:random} gives uncurated generated samples, which possess visually-pleasing quality and diversity.
We also explore the interpolation of the latent condition $\bm{z}$ between two generated avatars, as shown in Figure~\ref{fig:interpolation-more}, where we observe consistent   interpolation results with smooth appearance transition. Figure~\ref{fig:interpolation-more} shows additional results of  creating 3D portraits from a single reference image.

\section{Societal Impact}
The Rodin model  aims to enable a low-cost, fast and customizable creation experience of 3D digital avatars that refer to the traditional avatars manually created by 3D artists, as opposed to photorealistic avatars. The reason for focusing on digital avatars is twofold. On the one hand, digital avatars are widely used in movies, games, the metaverse, and the 3D industry in general. On the other hand, the available digital avatar data is very scarce as each avatar has to be painstakingly created by a specialized 3D artist using a sophisticated creation pipeline, especially for modeling hair and facial hair.  

Rather than collecting real photos, all our training images are rendered by Blender. Such synthetic data can mitigate the privacy and copyright concerns that existed in real face collection. Another advantage of using synthetic data is that we could have control over the variation and diversity of rendered images, eliminating the data  bias in existing face datasets. Also, digital avatars are easier to be distinguished from real people compared with photo-realistic avatars, hindering misuse for impersonating real persons. Nonetheless, the 3D portrait reconstruction and text-based avatar customization could still be misused for spreading disinformation maliciously, like all other AI-based content generation models. We caution that the high-quality renderings produced by our model may potentially be misused and viable solutions so avoid this include adding tags or watermarks when distributing the generated photos. 

This work successfully generalizes the power of diffusion models from 2D to 3D and is promising to offer the new design tool for 3D artists which could significantly save the costs of the traditional 3D modeling and rendering pipeline. In the next we intend to explore the possibility of modeling general 3D scenes using the same technique and investigate novel applications such as Lego and architect designs.


\end{document}